\RequirePackage{fix-cm}
\documentclass[smallcondensed]{svjour3}     
\smartqed  
\usepackage{graphicx}
\journalname{XXX}
\usepackage{hyperref}       
\usepackage{url}            
\usepackage{booktabs}       
\usepackage{amsfonts}       
\usepackage{nicefrac}       
\usepackage{microtype}      
\usepackage{amsmath}
\usepackage{mathrsfs}
\usepackage{color}
\usepackage[inline]{enumitem}

\usepackage{color}
\usepackage{cite}
\usepackage{url}            
\usepackage{amsfonts}       
\usepackage{amsfonts,amssymb}
\usepackage{algorithm}
\usepackage{algorithmic}

\DeclareMathOperator*{\argmin}{argmin}
\DeclareMathOperator*{\argmax}{argmax}

\newfloat{figtab}{htb}{fgtb}
\makeatletter
\newcommand\figcaption{\def\@captype{figure}\caption}
\newcommand\tabcaption{\def\@captype{table}\caption}
\makeatother

\begin{document}

\title{Tensor Q-Rank: New Data Dependent Definition of Tensor Rank
}

\author{Hao Kong         \and
        Canyi Lu  \and
        Zhouchen Lin
}


\institute{Hao Kong \at
              Key Lab. of Machine Perception (MOE), School of EECS, Peking University, Beijing, China.\\
              \email{konghao@pku.edu.cn}  
           \and
           Canyi Lu \at
           Department of Electrical \& Computer Engineering (ECE), Carnegie Mellon University, Pittsburgh, America.\\
           \email{canyilu@gmail.com}            
           \and
           Zhouchen Lin \at
              Key Lab. of Machine Perception (MOE), School of EECS, Peking University, Beijing, China.\\
              Z. Lin is the corresponding author.\\
              \email{zlin@pku.edu.cn}      
}

\date{Received: date / Accepted: date}

\maketitle

\begin{abstract}
Recently, the \textit{Tensor Nuclear Norm~(TNN)} regularization based on t-SVD has been widely used in various low tubal-rank tensor recovery tasks. However, these models usually require smooth change of data along the third dimension to ensure their low rank structures. 
In this paper, we propose a new definition of data dependent tensor rank named \textit{tensor Q-rank} by a learnable orthogonal matrix $\mathbf{Q}$, and further introduce a unified data dependent low rank tensor recovery model.
According to the low rank hypothesis, we introduce two explainable selection method of $\mathbf{Q}$, under which the data tensor may have a more significant low tensor Q-rank structure than that of low tubal-rank structure. Specifically, maximizing the variance of singular value distribution leads to Variance Maximization Tensor Q-Nuclear norm~(VMTQN), while minimizing the value of nuclear norm through manifold optimization leads to Manifold Optimization Tensor Q-Nuclear norm~(MOTQN).
Moreover, we apply these two models to the low rank tensor completion problem, and then give an effective algorithm and briefly analyze why our method works better than TNN based methods in the case of complex data with low sampling rate. Finally, experimental results on real-world datasets demonstrate the superiority of our proposed model in the tensor completion problem with respect to other tensor rank regularization models.

\end{abstract}

\section{Introduction}

With the development of data science, multi-dimensional data structures are becoming more and more complex. The low-rank tensor recovery problem, which aims to recover a low-rank tensor from an observed tensor, has also been extensively studied and applied. 
The problem can be formulated as the following model:
\begin{equation}\label{LRTC}
\min_\mathcal{X}\ \text{rank} (\mathcal{X}),\quad s.t.\ \Psi(\mathcal{X}) = \mathcal{Y},
\end{equation}
where $\mathcal{Y}$ is the observed measurement by a linear operator $\Psi(\cdot)$ and $\mathcal{X}$ is the clean data. 
Generally, it is difficult to solve Eq.~(\ref{LRTC}) directly, and different rank definitions correspond to different models. 
The commonly used definitions of tensor rank are all related to particular tensor decompositions~\cite{excel_9}. For example, CP-rank~\cite{excel_28} is based on the CANDECOMP/PARAFAC decomposition~\cite{excel_29}; multilinear rank~\cite{multilinear_rank} is based on the Tucker decomposition~\cite{excel_30}; tensor multi-rank and tubal-rank~\cite{excel_18} are based on t-SVD~\cite{excel_21}; and a new tensor rank with invertible linear operator~\cite{lu2019cvpr} is based on T-SVD~\cite{kernfeld2015tensor}. Among them, CP-rank and multilinear rank are both older and more widely studied, while the remaining two mentioned here are relatively new. Minimizing the rank function in Eq.~(\ref{LRTC}) directly is usually NP-hard and is difficult to be solved within polynomial time, hence we often replace $\text{rank}(\mathcal{\mathcal{X}})$ by a convex/non-convex surrogate function.
Similar to the matrix case~\cite{excel_7_1,excel_7_2}, with different definitions of tensor singular values, various tensor nuclear norms are proposed as the rank surrogates~\cite{excel_1_1,excel_12,excel_21,lu2019cvpr}. 

\subsection{Existing Mainstream Methods and Their Limitations}\label{Sec:1_1}
Friedland and Lim~\cite{excel_12} introduce cTNN (Tensor Nuclear Norm based on CP) as the convex relaxation of the tensor CP-rank:
\begin{equation}\label{cTNN}
\|\mathcal{T}\|_{cTNN} = \inf\left\{ \sum_{i=1}^{r} |\lambda_i|:\mathcal{T} = \sum_{i=1}^{r}\lambda_i \mathbf{u}_i\circ\mathbf{v}_i\circ\mathbf{w}_i \right\},
\end{equation}
where $\|\mathbf{u}_i\| = \|\mathbf{v}_i\| = \|\mathbf{w}_i\| = 1$ and $\circ$ represents the vector outer product\footnote{{Please see}~\cite{excel_9} or our supplementary materials for more details.}. 
However, for a given tensor $\mathcal{T}\in\mathbb{R}^{n_1\times n_2\times n_3}$, minimizing the surrogate objection $\|\mathcal{T}\|_{cTNN}$ directly is difficult due to the fact that computing CP-rank is usually NP-complete~\cite{excel_11,excel_42} and computing cTNN is NP-hard in some sense~\cite{excel_12}, which also mean we cannot verify the consistency of cTNN's implicit decomposition with the ground-truth CP-decomposition. 
{Meanwhile, it is hard to measure the cTNN's tightness relative to the CP-rank\footnote{For the matrix case, the nuclear norm is the conjugate of the conjugate function of the rank function in the unit ball. However, it is still unknown whether this property holds for cTNN and CP-rank.}. }
Although Yuan and Zhang~\cite{excel_13} give the sub-gradient of cTNN by leveraging its dual property, the high computational cost makes it difficult to implement.

To reduce the computation cost of computing the rank surrogate function, Liu et al.~\cite{excel_1_1} define a kind of tensor nuclear norm named SNN (Sum of Nuclear Norm) based on the Tucker decomposition~\cite{excel_30}:
\begin{equation}\label{intro_SNN}
\|\mathcal{T}\|_{SNN} = \sum_{i=1}^{{\color{blue}d}} \left\| \mathbf{T}_{(i)} \right\|_*,
\end{equation}
where $\mathcal{T} \in \mathbb{R}^{n_1\times \ldots\times n_{{\color{blue}d}}}$, $\mathbf{T}_{(i)}\in\mathbb{R}^{(n_1\ldots n_{i-1}n_{i+1}\ldots n_{{\color{blue}d}})\times n_i}$ denotes unfolding the tensor along the $i$-th dimension, and $\|\cdot\|_*$ is the nuclear norm of a matrix, i.e., sum of singular values. The convenient calculation algorithm makes SNN widely used~\cite{TNNLS_fu2016tensor,TNNLS_LiuGeneralized,excel_1_1,excel_15,excel_17_1}.
It is worth to mentioned that, although SNN has a similar representation to matrix case, Paredes and Pontil~\cite{excel_16} point out that SNN is not the tightest convex relaxation of the multilinear rank~\cite{multilinear_rank}, and is actually an overlap regularization of it. References~\cite{Latent_2010,Latent_2013,Latent_2014} also propose a new regularizer named Latent Trace Norm to better approximate the tensor rank function. 
In addition, due to unfolding the tensor directly along each dimension, the information utilization of SNN based model is insufficient.

To avoid information loss in SNN, Kilmer and Martin~\cite{excel_21} propose a tensor decomposition named t-SVD with a Fourier transform matrix $\mathbf{F}$, and Zhang et al.~\cite{excel_10} give a definition of the tensor nuclear norm on $\mathcal{T}\in\mathbb{R}^{n_1\times n_2 \times n_3}$ corresponding to t-SVD, i.e., Tensor Nuclear Norm~(TNN):
\begin{equation}\label{DefofTNN}
\|\mathcal{T}\|_{TNN} := \frac{1}{n_3}\sum_{i=1}^{n_3} \left\| \mathbf{G}^{(i)}\right\|_* , \quad \text{where }\  \mathcal{G} = \mathcal{T}\times_3 \mathbf{F},
\end{equation}
where $\mathbf{G}^{(i)}$ denotes the $i$-th frontal slice matrix of tensor $\mathcal{G}$\footnote{The implementation of Fourier transform along the third dimension of $\mathcal{T}$ is equivalent to multiplying a DFT matrix $\mathbf{F}$ by using $\times_3$. For more details, {please see} Sec.~\ref{Sec:TQR}.}, and $\times_3$ is the mode-$3$ multilinear multiplication~\cite{excel_30}.
{Benefitting from the efficient Discrete Fourier transform and the better sampling effect of Fourier basis on time series features,}
TNN has attracted extensive attention in recent years~\cite{excel_10,excel_1_2,LuIJCAI2018,TNNLS_yin2018multiview,TNNLS_hu2016twist}. The operation of Fourier transform along the third dimension makes TNN based models have a natural computing advantage for video and other data with strong time continuity along a certain dimension. 

However, when considering the smoothness of different data, using a fixed Fourier transform matrix $\mathbf{F}$ may bring some limitations. In this paper, we define smooth and non-smooth data along a certain dimension as the usual intuitive meaning, which means the slices of tensor data along a dimension are arranged in a certain paradigm, e.g., time series. For example, a continuous video data is smooth. But if the data tensor is a concatenation of several different scene videos or a random arrangement of all frames, then the data is non-smooth.

Firstly, TNN needs to implement Singular Value Decomposition~(SVD) in the complex field~$\mathbb{C}$, which is slightly slower than that in the real field~$\mathbb{R}$. 
Besides, the experiments in related papers~\cite{excel_10,LuIJCAI2018,excel_22,Kong2018} are usually based on some special dataset which have smooth change along the third dimension, such as RGB images and short videos. 
Those non-smooth data may increase the number of non-zero tensor singular values~\cite{excel_21,excel_10}, weakening the significance of low rank structure.
Since tensor multi-rank~\cite{excel_10} is actually the rank of each projection matrix on different Fourier basis, the non-smooth change along the third dimension may lead to large singular values appearing on the projection matrix slices which are corresponding to the high frequency. 

\subsection{Related Work}

\begin{figure}[t]
	\centering
	\includegraphics[width=0.6\columnwidth]{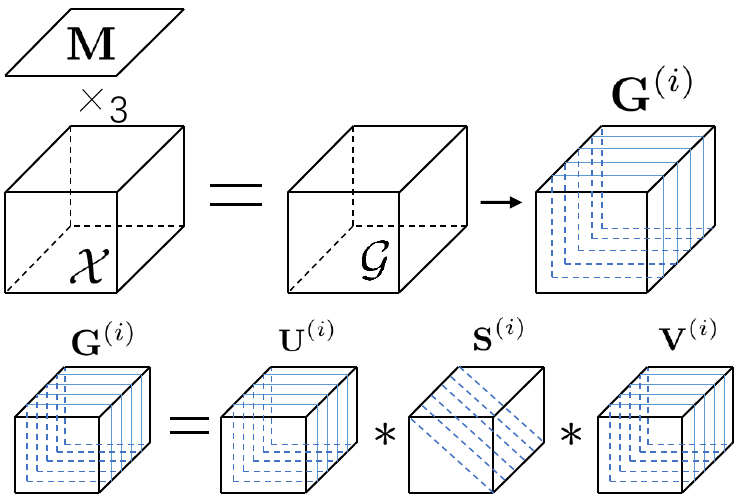}
	\caption{Replace $\mathbf{F}$ in Eq.~(\ref{DefofTNN}) by matrix $\mathbf{M}$ and further obtain new definitions of tensor rank $\operatorname{rank}_M(\mathcal{X})$ and tensor nuclear norm $\|\mathcal{X}\|_{M,*}$ by using $\mathbf{S}^{(i)}$.}\label{intro_show_product}
\end{figure}

In order to solve the above phenomenon, {there are some works~\cite{kernfeld2015tensor,xu2019fast,song2019robust,lu2019cvpr,jiang2020framelet} that consider to improve the projection matrix of TNN,} i.e., the Discrete Fourier transform matrix $\mathbf{F}$ in Eq.~(\ref{DefofTNN}). 
These work want to replace $\mathbf{F}$ by another measurement matrix $\mathbf{M}$ and further obtain new definitions of tensor rank $\operatorname{rank}_M(\mathcal{X})$ and tensor nuclear norm $\|\mathcal{X}\|_{M,*}$ as regularizers. Figure~\ref{intro_show_product} shows the related operations. Their recovery models can be summarized as follows:
\begin{equation}\label{intro_M_model}
\min_{\mathcal{X}}\ \left\| \mathcal{X} \right\|_{M,*},\quad s.t.\ \Psi(\mathcal{X}) = \mathcal{Y},\  \mathbf{M} \text{ is determined by some prior knowledge}.
\end{equation}
{Please see} Sec.~\ref{notations_and_preliminaries} for the relevent definitions in Eq.~(\ref{intro_M_model}). In the following, we will discuss the motivations and limitations of these work~\cite{kernfeld2015tensor,xu2019fast,song2019robust,lu2019cvpr,jiang2020framelet}, respectively.

Kernfeld, Kilmer, and Aeron~\cite{kernfeld2015tensor} generalize the t-product by introducing a new operator named cosine transform product with an \textbf{arbitrary invertible} linear transform $\mathcal{L}$~(or \textbf{arbitrary invertible} matrix $\mathbf{M}$). 
For a given $\mathcal{T}\in\mathbb{R}^{n_1\times n_2\times n_3}$ and an invertible matrix $\mathbf{M}\in\mathbb{R}^{n_3\times n_3}$, they have $\mathcal{L}_{\mathbf{M}}(\mathcal{T}) = \mathcal{T}\times_3 \mathbf{M}$ and $\mathcal{L}_{\mathbf{M}}^{-1}(\mathcal{T}) = \mathcal{T}\times_3 \mathbf{M}^{-1}$. 
Different from the commonly used definition of tensor mode-$i$ product in~\cite{excel_9,excel_1_1,kernfeld2015tensor,lu2019cvpr}, it should be mentioned that for convenience in this paper, we define $\mathcal{L}_{\mathbf{Q}}(\mathcal{T}) =\mathcal{T}\times_3 \mathbf{Q} = \operatorname{fold_3}(\mathbf{T}_{(3)}\mathbf{Q})$, where $\mathbf{T}_{(3)} \in \mathbb{R}^{n_1 n_2\times n_3}$ and is defined by $\mathbf{T}_{(3)} := \operatorname{unfold_3}(\mathcal{T})$. That is to say, {we arrange the tensor fiber $\mathcal{T}_{ij:}$ by rows.}

Following this idea, Lu, Peng, and Wei~\cite{lu2019cvpr} propose a new tensor nuclear norm induced by invertible linear transforms~\cite{kernfeld2015tensor}. Different from~\cite{excel_21,excel_10}, they use an fixed invertible matrix to replace the Fourier transform matrix in TNN. Although this method improves the performance of the recovery model to a certain extent, some new problems still arise, such as how to determine the fixed invertible matrix. Normally, different data need different optimal invertible matrix, but a reasonable matrix selection method is not given in~\cite{lu2019cvpr}. Furthermore, the Frobenius norm of the invertible matrix is uncertain, which may lead to some computational problems, e.g., approaching zero or infinity. 

Additionally, Kernfeld, Kilmer, and Aeron~\cite{kernfeld2015tensor} propose an idea that, with the help of Toeplitz-plus-Hankel matrix~\cite{ng1999fast}, the Discrete cosine transform matrix $\mathbf{C}$ can also be used to replace $\mathbf{F}$. Then the work~\cite{xu2019fast} propose some fast algorithms for diagonalization and the relevant recovery model. However, $\mathbf{C}$ is still based on trigonometric function, and may lead to the similar problems with TNN based model, as we mentioned in the last paragraph of Sec.~\ref{Sec:1_1}.

Considering the efficiency of time-space transformation, the work~\cite{song2019robust} use the Daubechies $4$ discrete wavelet transform matrix to replace $\mathbf{F}$. As we know, the wavelet transform can take the position information into account, which may make it better than Fourier transform and cosine transform in handling some special data, e.g., audio data. However, many wavelet bases generate transform matrices in exponential form, which means the large scale wavelet matrix may bring the problem of computational complexity.

Regardless of the computational complexity, Jiang et al.~\cite{jiang2020framelet} introduce a new projection matrix named tight framelets transform matrix~\cite{cai2008framelet,jiang2018matrix}. They claim that redundancy in the transformation is important as such transformed coefficients can contain information of missing data in the original
domain~\cite{cai2008framelet}. However, we consider that redundancy is not a sufficient condition to improve the effect of recovery model shown in Eq.~(\ref{intro_M_model}).

In summary, different multipliers $\mathbf{M}$ in Eq.~(\ref{intro_M_model}) lead to different definitions of regularizer, which may lead to different experimental results. However, there is still no unified rules for selecting $\mathbf{M}$. 
It can be seen from the above methods that when $\mathbf{M}$ is selected as orthogonal matrix, it is convenient for calculation and interpretation. In general, projection bases are unit orthogonal. We further think that every data should have its best matching matrix, i.e., $\mathbf{M}$ could be data dependent.
In this paper, we solve the problem of how to define a better data dependent orthogonal transformation matrix.

\subsection{Motivation}

In the tensor completion task, we find that when dealing with some non-smooth data, Tensor Nuclear Norm~(TNN) based methods usually perform worse than the cases with smooth data. Therefore, we want to improve this phenomenon by changing the projection basis $\mathbf{F}$ in Eq.~(\ref{DefofTNN}). In other words, we provide some interpretable selection criteria of $\mathbf{M}$ in Eq.~(\ref{intro_M_model}), e.g., make $\mathbf{M}$ be an orthogonal matrix and data dependent w.r.t. the data tensor $\mathcal{X}$. The following gives the details:
\begin{equation}\label{motivation_q_recovery_model}
\min_{\mathcal{X},\mathbf{Q}}\ \left\| \mathcal{X} \right\|_{Q,*},\quad s.t.\ \Psi(\mathcal{X}) = \mathcal{Y},\  \ \mathbf{Q}^\top\mathbf{Q} = \mathbf{I},\ \mathbf{Q} \text{ is determined by } \mathcal{X}.
\end{equation}

\begin{figure}[t]
	\centering
	\includegraphics[width=0.4\columnwidth]{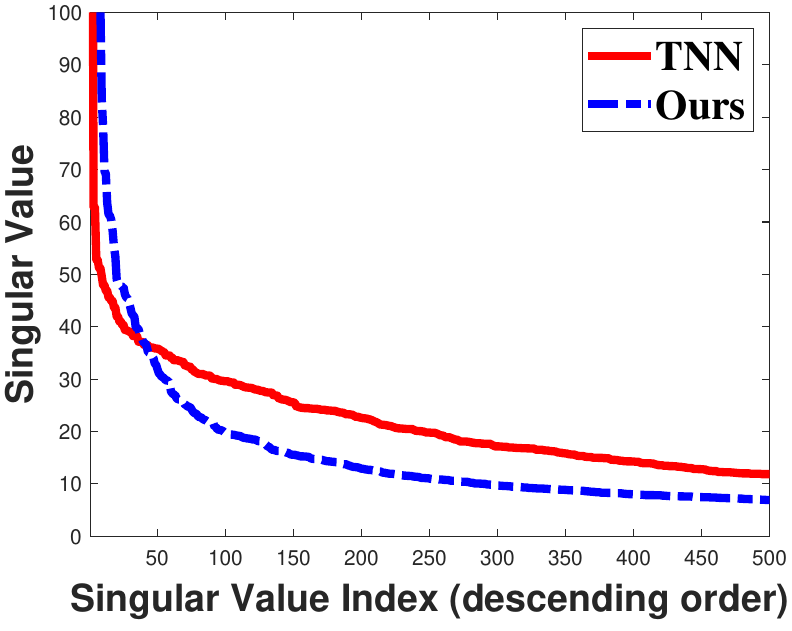}
	\ 
	\includegraphics[width=0.35\columnwidth]{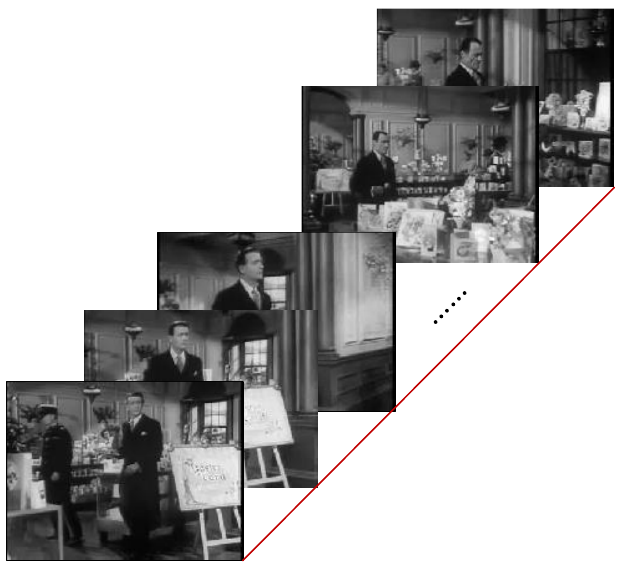}
	\caption{Compare the two different low rank structures between our proposed regularization and TNN regularization in non-smooth video data. \textbf{Left:} the first $500$ sorted singular values by TNN regularization~(divided by $\sqrt{n_3}$) and ours. \textbf{Right:} the short video with background changes.}\label{Intro}
\end{figure}

Whether in the case of matrix recovery~\cite{excel_7_1,excel_7_2} or tensor recovery~\cite{excel_20,LuIJCAI2018,lu2019cvpr}, the low rank hypothesis is very important. Generally speaking, the lower the rank of the data, the easier it is to recover with fewer observations. As can be seen from Figure.~\ref{Intro},  we can use a better $\mathbf{Q}$ to make the low rank structure of the non-smooth data more significant.

Considering the convex relaxation, the low rank property is usually represented by \textbf{(a): the distribution of singular values}, or \textbf{(b): the value of nuclear norm}. We may as well take these two points as priori knowledge respectively, and specify the selection rules of $\mathbf{Q}$ in Eq.~(\ref{motivation_q_recovery_model}), so that the low rank property of $\mathcal{X}$ can be better reflected. Therefore, we provide two methods in this paper as follows:

(a): {Let $\mathbf{Q}$ satisfy a certain selection method to make more tensor singular values close to $0$ while the remaining ones are far from $0$.} From another perspective, the distribution variance of singular values should be larger, which leads to Variance Maximization Tensor Q-Nuclear norm~(VMTQN) in Sec.~\ref{Sec:VMTQN}.

(b): Let $\mathbf{Q}$ minimize the nuclear norm $\left\| \mathcal{X} \right\|_{Q,*}$ directly, leading to a bilevel problem. As we know, nuclear norm is usually used as an surrogate function of the rank function. Then we use some manifold optimization method to solve the problem, which leads to Manifold Optimization Tensor Q-Nuclear norm~(MOTQN) in Sec.~\ref{Sec:MOTQN}.

\subsection{Contributions}
In summary, our main contributions include:
\begin{itemize}
	\item We propose a unified data dependent low rank tensor recovery model which is shown in Eq.~(\ref{motivation_q_recovery_model}). Among them, the corresponding definitions of tensor Q-rank $\operatorname{rank}_Q (\mathcal{X})$ and tensor Q-nuclear norm $\left\| \mathcal{X} \right\|_{Q,*}$ are proposed along with the learnable data dependent orthogonal $\mathbf{Q}$.
	\item From the low rank hypothesis, we consider the distribution of singular values and the value of nuclear norm as prior knowledge respectively, leading to two different selection rules of $\mathbf{Q}$. 
	It should be noted that both methods are designed to make the low rank structure more significant. Figure.~\ref{Intro} shows an example with background changing video data that, under our proposed selection of $\mathbf{Q}$, our low rank structure is more significant.
	\item For each method, we give relatively complete theoretical derivations, including interpretation and optimization. As for VMTQN in Sec.~\ref{Sec:VMTQN}, we start from variance maximization and use Theorem~\ref{Theorem_From_Lemma_2} to associate $\ell_{2,1}$ norm minimization with singular value decomposition, {and further make $\mathbf{Q}$ select as the matrix of right singular vectors.} {On the other hand, MOTQN in Sec.~\ref{Sec:MOTQN} minimizes the nuclear norm directly and use manifold optimization algorithm to update $\mathbf{Q}$ in each iteration. }
	\item Finally, we apply our proposed regularizers with adaptive $\mathbf{Q}$ to the tensor completion problem. We analyze the computational complexity, convergence and performance guarantee of our algorithm to a certain extent. Moreover, we explain why the more significant the low rank structure, the easier the data can be recovered, which corresponds to our motivation.
\end{itemize}

\section{Notations and Preliminaries}\label{notations_and_preliminaries}

\subsection{Notations}

We introduce some notations and necessary definitions which will be used later. 
Tensors are represented by uppercase calligraphic letters, e.g., $\mathcal{T}$.
Matrices are represented by boldface uppercase letters, e.g., $\mathbf{M}$.
Vectors are represented by boldface lowercase letters, e.g., $\mathbf{v}$.
Scalars are represented by lowercase letters, e.g., $s$.
{Given a third-order tensor $\mathcal{T}\in \mathbb{R}^{n_1\times n_2\times n_3}$,} we use $\mathbf{T}^{(k)}$ to represent its $k$-th frontal slice $\mathcal{T}(:,:,k)$ while its $(i,j,k)$-th entry is represented as $\mathcal{T}_{ijk}$.
$\sigma_i(\mathbf{X})$ denotes the $i$-th largest singular value of matrix $\mathbf{X}$. 
$\mathbf{X}^{+}$ denotes the pseudo-inverse matrix of $\mathbf{X}$. $\|\mathbf{X}\|_\sigma = \sigma_1(\mathbf{X})$ denotes the matrix spectral norm. $\|\mathbf{X}\|_* = \sum_{i=1}^{\min\{n_1,n_2\}} \sigma_i(\mathbf{X})$ denotes the matrix nuclear norm and $\|\mathbf{X}\|_{2,1} = \sum_{j=1}^{n_2}\sqrt{\sum_{i=1}^{n_1}\mathbf{X}_{ij}^2}$ denotes the matrix $\ell_{2,1}$ norm, where $\mathbf{X}\in\mathbb{R}^{n_1 \times n_2}$ and $\mathbf{X}_{ij}$ is the $(i,j)$-th entry of $\mathbf{X}$.

$\mathbf{T}_{(3)}\in\mathbb{R}^{n_1 n_2\times n_3}$ denotes unfolding the tensor $\mathcal{T}$ along the $3$-th dimension by columns, which is little different from~\cite{excel_9,kernfeld2015tensor}. That is to say, we arrange the tensor fiber $\mathcal{T}_{ij:}$ by columns. We then define $\mathcal{L}_{\mathbf{Q}}(\mathcal{T}) =\mathcal{T}\times_3 \mathbf{Q} = \operatorname{fold_3}(\mathbf{T}_{(3)}\mathbf{Q})$ and have $\mathcal{L}_{\mathbf{Q}}^{-1}(\mathcal{T}) = \mathcal{T}\times_3 \mathbf{Q}^{-1}$,
where $\mathbf{T}_{(3)} \in \mathbb{R}^{n_1 n_2\times n_3}$ and is defined by $\mathbf{T}_{(3)} := \operatorname{unfold_3}(\mathcal{T})$. 
Due to limited space, for the definitions of $\mathcal{P}_{\mathcal{T}}$~\cite{excel_1_2}, multilinear multiplication~\cite{excel_30}, t-product~\cite{excel_21}, and so on, {please see} our Supplementary Materials.

\subsection{Tensor Q-rank}\label{Sec:TQR}
For a given tensor $\mathcal{X}\in\mathbb{R}^{n_1\times n_2\times n_3}$ and a Fourier transform matrix $\mathbf{F}\in\mathbb{C}^{n_3\times n_3}$, if we use $\mathbf{G}^{(i)}$ to represent the $i$-th frontal slice of tensor $\mathcal{G}$, then the tensor multi-rank and Tensor Nuclear Norm~(TNN) of $\mathcal{X}$ can be formulated by mode-$3$ multilinear multiplication as follows:
\begin{eqnarray}
\operatorname{rank_m} :=&\left\{(r_1,\ldots,r_{n_3})\  \big|\  r_i = \operatorname{rank}\left( \mathbf{G}^{(i)}\right), \mathcal{G} = \mathcal{X}\times_3 \mathbf{F} \right\},\label{Eq:tensor_multi_rank}\\
\|\mathcal{X}\|_* :=& \frac{1}{n_3}\sum_{i=1}^{n_3} \left\| \mathbf{G}^{(i)}\right\|_* , \quad \text{where }\  \mathcal{G} = \mathcal{X}\times_3 \mathbf{F}.\label{Def_of_TNN}
\end{eqnarray}

Comparing with CP-rank and cTNN mentioned in Sec.~\ref{Sec:1_1}, it is quite easy to calculate Eqs.~(\ref{Eq:tensor_multi_rank}) and~(\ref{Def_of_TNN}) through the matrix Singular Value Decomposition~(SVD). Kernfeld, Kilmer, and Aeron~\cite{kernfeld2015tensor} generalize the t-product by introducing a new operator named cosine transform product with an {arbitrary invertible} linear transform $\mathcal{L}$~(or {arbitrary invertible} matrix $\mathbf{Q}$). 
For an invertible matrix $\mathbf{Q}\in\mathbb{R}^{n_3\times n_3}$, they have $\mathcal{L}_{\mathbf{Q}}(\mathcal{X}) = \mathcal{X}\times_3 \mathbf{Q}$ and $\mathcal{L}_{\mathbf{Q}}^{-1}(\mathcal{X}) = \mathcal{X}\times_3 \mathbf{Q}^{-1}$. 

Here, we further define the invertible multiplier $\mathbf{Q}$ as any general real orthogonal matrix. It is worth mentioning that the orthogonal matrix $\mathbf{Q}$ has two good properties: {one is invertibility,} the other is to keep Frobenius norm invariant, i.e., $\|\mathcal{X}\|_F = \|\mathcal{L}_{\mathbf{Q}}(\mathcal{X})\|_F$. Then we introduce a new definition of tensor rank named Tensor Q-rank.

\begin{definition}\textbf{(Tensor Q-rank)}
	Given a tensor $\mathcal{X}\in\mathbb{R}^{n_1\times n_2\times n_3}$ and a {fixed} real orthogonal matrix $\mathbf{Q} \in \mathbb{R}^{n_3\times n_3}$, the tensor Q-rank of $\mathcal{T}$ is defined as the following:
	\begin{equation}\label{tensor_q_rank_cal_tucker_product}
	\operatorname{rank}_Q (\mathcal{X}) := \sum_{i=1}^{n_3} \operatorname{rank}\left( \mathbf{G}^{(i)}\right) ,  \text{where }\  \mathcal{G} = \mathcal{L}_{\mathbf{Q}}(\mathcal{X}) = \mathcal{T}\times_3 \mathbf{Q}.
	\end{equation}
	The corresponding low rank tensor recovery model can be written as follows:
	\begin{equation}\label{low_tensor_q_rank_model_with_rank}
	\min_\mathcal{X}\ \text{rank}_Q (\mathcal{X}),\quad s.t.\ \Psi(\mathcal{X}) = \mathcal{Y}.
	\end{equation}
\end{definition}

Generally in the low rank recovery models, due to the discontinuity and non-convexity of the rank function, it is quite difficult to minimize the rank function directly.
Therefore, some auxiliary definitions of tensor singular value and tensor norm are needed to relax the rank function. 

\subsection{Definitions of Tensor Singular Value and Tensor Norm}
Considering the superior recovery performance of TNN in many existing tasks, e.g., video denoising~\cite{lu2019tensorpami} and subspace clustering~\cite{TNNLS_yin2018multiview}, we can use the similar singular value definition of TNN. Given a tensor $\mathcal{X}\in\mathbb{R}^{n_1\times n_2\times n_3}$ and a fixed orthogonal matrix $\mathbf{Q}$ such that $\mathcal{G} = \mathcal{L}_{\mathbf{Q}}(\mathcal{X})$, then the $\mathbf{Q}$-singular value of $\mathcal{X}$ is defined as $\{\sigma_j(\mathbf{G}^{(i)}) \}$, where $i=1,\ldots,n_3$, $j=1,\ldots,\min\{n_1,n_2\}$, $\mathbf{G}^{(i)}$ is the $i$-the frontal slice of $\mathcal{G}$, and $\sigma(\cdot)$ denotes the matrix singular value.
When an orthogonal matrix $\mathbf{Q}$ is fixed, the corresponding tensor spectral norm and tensor nuclear norm of $\mathcal{X}$ can also be given.

\begin{definition}\textbf{(Tensor Q-spectral norm and Tensor Q-nuclear norm)}
	Given a tensor $\mathcal{X}\in\mathbb{R}^{n_1\times n_2\times n_3}$ and a fixed real orthogonal matrix $\mathbf{Q} \in \mathbb{R}^{n_3\times n_3}$, the tensor Q-spectral norm and tensor Q-nuclear norm of $\mathcal{X}$ are defined as the followings:
	\begin{equation}\label{q_spectral}
	\left\| \mathcal{X} \right\|_{Q,\sigma} := \max_i \left\lbrace \left\| \mathbf{G}^{(i)} \right\|_\sigma \Big|\  \mathcal{G} = \mathcal{L}_{\mathbf{Q}}(\mathcal{X})  \right\rbrace .
	\end{equation}
	\begin{equation}\label{q_nuclear}
	\left\| \mathcal{X} \right\|_{Q,*} := \sum_{i=1}^{n_3} \left\| \mathbf{G}^{(i)}\right\|_* , \quad \text{where }\  \mathcal{G} = \mathcal{L}_{\mathbf{Q}}(\mathcal{X}).
	\end{equation}
\end{definition}

Moreover, with any fixed orthogonal matrix $\mathbf{Q}$, the convexity, duality, and envelope properties are all preserved. 

\begin{property}\textbf{(Convexity)}\label{Property_1}
	Tensor Q-nuclear norm and Tensor Q-spectral norm are both convex.
\end{property}

\begin{property}\textbf{(Duality)}\label{Property_2}
	Tensor Q-nuclear norm is the dual norm of Tensor Q-spectral norm, and vice versa.
\end{property}

\begin{property}\textbf{(Convex Envelope)}\label{Property_3}
	Tensor Q-nuclear norm is the tightest convex envelope of the Tensor Q-rank within the unit ball of the Tensor Q-spectral norm.
\end{property}
These three properties are quite important in the low rank recovery theory. Property~\ref{Property_3} implies that we can use the tensor Q-nuclear norm as a rank surrogate. That is to say, when the orthogonal matrix $\mathbf{Q}$ is given, we can replace the low tensor Q-rank model~(\ref{low_tensor_q_rank_model_with_rank}) with model~(\ref{low_tensor_q_rank_model_with_norm}) to recover the original tensor:
\begin{equation}\label{low_tensor_q_rank_model_with_norm}
\min_\mathcal{X}\ \left\| \mathcal{X} \right\|_{Q,*},\quad s.t.\ \Psi(\mathcal{X}) = \mathcal{Y}.
\end{equation}

In some cases, we will encounter the case that $\mathbf{Q}$ is not a square matrix, i.e., $\mathbf{Q}\in\mathbb{R}^{n_3\times r}$ is column orthonormal. Then the corresponding definitions of $\operatorname{rank}_Q(\mathcal{X})$ in Eq.~(\ref{tensor_q_rank_cal_tucker_product}) and $\left\| \mathcal{X} \right\|_{Q,*}$ in Eq.~(\ref{q_nuclear}) also change to the sum of $r$ frontal slices instead of $n_3$.
Moreover, as for the convex envelope property, the double conjugate function of rank function $\operatorname{rank}_Q(\mathcal{X})$ is still the corresponding nuclear norm $\left\| \mathcal{X} \right\|_{Q,*}$ within an unit ball. We give the following theorem to illustrate this case:
\begin{theorem}\label{non_square_theorem}
	Given a tensor $\mathcal{T}\in\mathbb{R}^{n_1\times n_2\times n_3}$ and a fixed real column orthonormal matrix $\mathbf{Q} \in \mathbb{R}^{n_3\times r}$. Let $\mathbf{Q}_\perp\in\mathbb{R}^{n_3\times (n_3-r)}$ be the column complement matrix of $\mathbf{Q}$, and 
	$\mathbf{Q}_t = \begin{bmatrix}
	\mathbf{Q} & \mathbf{Q}_\perp
	\end{bmatrix}$ be a orthogonal matrix. Then within the unit ball $\mathcal{D} = \{ \mathcal{X} | \|\mathcal{X}\|_{{Q}_t} \leq 1 \}$, the double conjugate function of $\operatorname{rank}_Q(\mathcal{X})$ is $\left\| \mathcal{X} \right\|_{Q,*}$:
	\begin{equation}
	rank_Q^{**}(\mathcal{X}) = \|\mathcal{X}\|_{Q,*}.
	\end{equation}
	In other words, $\|\mathcal{X}\|_{Q,*}$ is still the tightest convex envelope of $rank_Q^{**}(\mathcal{X})$ within the unit ball $\mathcal{D}$.
\end{theorem}
Theorem~\ref{non_square_theorem} indicate that even if $Q$ is not a square matrix, Eq.~(\ref{low_tensor_q_rank_model_with_norm}) can still be used as an effective recovery model.

\section{Two Ways for Determining $\mathbf{Q}$: Maximizing Variance \& Stiefel Manifold Optimization}

In practical problems, the selection of $\mathbf{Q}$ often has a tremendous impact on the performance of the model~(\ref{low_tensor_q_rank_model_with_norm}). If $\mathbf{Q}$ is an identity matrix $\mathbf{I}$, it is equivalent to solving each frontal slice separately by the low rank matrix methods~\cite{excel_7_1}. Or if $\mathbf{Q}$ is a Fourier transform matrix $\mathbf{F}$, it is equivalent to the TNN-based methods~\cite{excel_10,excel_1_2,excel_20}. 
Through the analysis of \cite{lu2019cvpr} and our previous section, for a given data $\mathcal{X}$, those $\mathbf{Q}$ that make $\operatorname{rank}_Q(\mathcal{X})$ lower usually make the recovery problem~(\ref{low_tensor_q_rank_model_with_norm}) easier.

Following, if we let $\mathbf{Q}$ in Eq.~(\ref{low_tensor_q_rank_model_with_rank}) and~(\ref{low_tensor_q_rank_model_with_norm}) be a learnable variable \textbf{w.r.t.} data tensor $\mathcal{X}$, we can get a \textbf{data-dependent} tensor rank and corresponding low rank recovery model:
\begin{equation}\label{bilevel_low_tensor_q_rank_model}
\min_{\mathcal{X},\mathbf{Q}}\ \left\| \mathcal{X} \right\|_{Q,*},\quad s.t.\ \Psi(\mathcal{X}) = \mathcal{Y},\  \mathbf{Q} \text{ is determined by } \mathcal{X}.
\end{equation}
Easy to see that Eq.~(\ref{bilevel_low_tensor_q_rank_model}) is actually a bilevel model and is usually hard to be solved directly. In the following, we will show two ways to solve this problem from the following two perspectives: 
\begin{enumerate}
	\item One is to use the prior knowledge of $\mathcal{X}$ to specify the selection criteria of $\mathbf{Q}$. For the low rank hypothesis, we usually measure it by the distribution of singular values. 
	Therefore, we consider artificially specifying the conditions that $\mathbf{Q}$ should  satisfy so as to maximize the variance of the corresponding singular values.
	\item The other is to give the function $\mathbf{Q} = \argmin f(\mathcal{X},\mathbf{Q}) = \argmin \|\mathcal{X}\|_{Q,*}$ and then use manifold optimization to solve the bilevel model directly. 
	That is to say, We directly minimize the surrogate function of rank function~(Property~\ref{Property_3} and Theorem~\ref{non_square_theorem}).
	It should be noted that although this way has higher rationality, it corresponds to a higher computational complexity.
\end{enumerate}
From the above two perspectives, $\mathbf{Q}$ will be data dependent. In the following, we will introduce our two methods in two sub-sections respectively~(\textbf{Sec.}\ref{Sec:VMTQN} and \textbf{Sec.}\ref{Sec:MOTQN}). And in the last part~(\textbf{Sec.}\ref{Sec:two_case}), considering a {third-order tensor} $\mathcal{X}\in\mathbb{R}^{n_1\times n_2\times n_3}$, we analyze the applicability of each method in two different situations, i.e., $n_1n_2 < n_3$ and $n_1n_2 > n_3$.

\subsection{\textbf{Way I~(VMTQN):} Specify the Selection of $\mathbf{Q}$ by Variance Maximization}\label{Sec:VMTQN}

Let $\mathcal{G} = \mathcal{L}_\mathbf{Q}(\mathcal{X}) = \mathcal{X}\times_3 \mathbf{Q}$ and $\{\mathbf{G}^{(i)}\}_i$ denotes the frontal slices of $\mathcal{G}$.
We hope to find a data-dependent $\mathcal{L}_\mathbf{Q}$ in Eqs.~(\ref{q_nuclear}) and~(\ref{low_tensor_q_rank_model_with_norm}) instead of $\mathcal{L}_\mathbf{F}$ in TNN~(Eq.~(\ref{Def_of_TNN})), which can reduce the number of non-zero singular values of each projected slices $\mathbf{G}^{(i)}$. Our analyses are as follows.
{
\newline
\textbf{(1):} If we make $\mathbf{Q}$ an orthogonal matrix, then it is also invertible. By using the unitary invariance of the Frobenius norm, the sum of the squares of each projected slice's Frobenius norm is a \textbf{constant $C$}, i.e., $\sum_{i=1}^{n_3} \|\mathbf{G}^{(i)}\|_F^2 = \|\mathcal{X}\|_F^2 = C$. Therefore, we need to consider how to select $\mathbf{Q}$ to make more singular values of $\{ \mathbf{G}^{(i)} \}$ close to zero while the square sum of all singular values is a constant, i.e., $\sum_{j=1}^{n_3}\sigma_j(\mathbf{G}^{(i)})^2 = C$.
\newline
\textbf{(2):} Considering the definitions of tensor rank, tensor norm and tensor singular value corresponding to TNN in~\cite{excel_10,excel_20}, and tensor Q-rank in this paper, the matrix inequality $\frac{1}{n}\sum_{j=1}^{n}\sigma_j(\mathbf{G}^{(i)}) \leq \|\mathbf{G}^{(i)}\|_\sigma \leq  \|\mathbf{G}^{(i)}\|_F$~(singular value, spectral norm and Frobenius norm, respectively) implies that, the closer $\|\cdot\|_F$ is to zero, the more singular values are close to zero, which will lead to a more significant tensor low rank structure~(w.r.t. $\operatorname{rank}_Q(\mathcal{X})$) with high probability. 
}

\subsubsection{From Variance Maximization to Singular Matrix}\label{Sec:From_VM_to_SM}

Combined with above two points, it is easy to see that we need to make more $\|\mathbf{G}^{(i)}\|_F$ close to $0$ while the sum of squares $\sum_{i=1}^{n_3} \|\mathbf{G}^{(i)}\|_F^2$ is a constant $C$. 
From the perspective of variable distribution, we need to choose a data-dependent $\mathbf{Q}$ to maximize the distribution \textbf{variance} of $\{ \|\mathbf{G}^{(i)}\|_F \}$, where $\mathcal{G} = \mathcal{L}_\mathbf{Q}(\mathcal{X})$ and $\mathbf{G}^{(i)}$ is the $i$-th frontal slice matrix of $\mathcal{G}$. For better explanations, we give the following two Lemmas, {and the optimality condition of Lemma~\ref{Intro_lemma_1} illustrate our hypothesis that there should be more $\|\mathbf{G}^{(i)}\|_F$ close to $0$}\footnote{Notice that minimizing $\sum_{i=1}^{n} a_i$ in Lemma~\ref{Intro_lemma_1} can be seen as a linear hyperplane optimization problem defined in the first quadrant Euclidean spherical surface: $\{(a_1,\ldots,a_n)|\sum_{i=1}^{n} a_i^2 = C, a_i\geq 0\}$. The intersection of sphere and each axis is distributed on the optimal hyperplane, which corresponds to only one non-zero coordinate~(more variables close to $0$). }.
\begin{lemma}\label{Intro_lemma_1}
	Given $n$ non-negative variables $\{a_1,a_2,\ldots,a_n\}$ such that $\sum_{i=1}^{n} a_i^2 = C$, then maximizing the variance $\text{Var}[a_i]$ is equivalent to minimizing the summation $\sum_{i=1}^{n} a_i$. Moreover, the optimal condition is that there is only one non-zero variable in $\{a_1,a_2,\ldots,a_n\}$. {Please see} Appendix~\ref{APP_1} for proof.
\end{lemma}

By using Lemma~\ref{Intro_lemma_1}, maximizing the variance of $\{ \|\mathbf{G}^{(i)}\|_F \}$ is equivalent to minimizing the sum $\sum_{i=1}^{n_3}\|\mathbf{G}^{(i)}\|_F$. Then we have $\sum_{i=1}^{n_3}\|\mathbf{G}^{(i)}\|_F = \|\mathbf{G}_{(3)}\|_{2,1} = \|\mathbf{X}_{(3)}\mathbf{Q}\|_{2,1}$, where $\mathbf{G}_{(3)}$ and $\mathbf{X}_{(3)}$ denote the mode-$3$ unfolding matrices~\cite{excel_30}.
\begin{lemma}\label{Intro_lemma_2}
	Given a fixed matrix $\mathbf{X}\in\mathbb{R}^{n_1\times n_2}$, and its full Singular Value Decomposition as $\mathbf{X} = \mathbf{U}\mathbf{\Sigma}\mathbf{V}^\top$ with $\mathbf{U}\in\mathbb{R}^{n_1\times n_1}$, $\mathbf{\Sigma}\in\mathbb{R}^{n_1\times n_2}$, and $\mathbf{V}\in\mathbb{R}^{n_2\times n_2}$. Then {the matrix of right singular vectors} $\mathbf{V}$ optimizes the following:
	\begin{equation}\label{Lemma_1_equation}
	\min_{\mathbf{Q}\in\mathbb{R}^{n_2 \times n_2}}\|\mathbf{X}\mathbf{Q}\|_{2,1},\quad \text{s.t.}\ \mathbf{Q}^\top\mathbf{Q} = \mathbf{I},
	\end{equation}
	where $\|\mathbf{M}\|_{2,1} = \sum_{i=1}^{col} \|\mathbf{M}_{:,i}\|_2$ is the sum of the $\ell_2$ norms of all column vectors. {Please see} Appendix~\ref{APP_2} for proof.
\end{lemma}

Lemma~\ref{Intro_lemma_1} turns the maximizing variance problem into minimizing summation problem, while Lemma~\ref{Intro_lemma_2} gives a feasible solution to the problem of minimizing the summation of $\ell_2$ norm.
However, when $n_1\leq n_2$, there will be some zero-columns appearing in $\mathbf{\Sigma}$. We can use skinny SVD to reduce the redundant columns of $\mathbf{Q}$ in Eq.~(\ref{Lemma_1_equation}).
Note that the size of $\mathbf{V}$ in skinny SVD is related to the size of $\mathbf{X}$.
Considering the two cases $n_1\geq n_2$ and $n_1 < n_2$ of $\mathbf{X}\in\mathbb{R}^{n_1 \times n_2}$, we introduce an auxiliary variable $r = \min\{ n_1,n_2\}$ to unify {the matrix of right singular vectors} as $\mathbf{V} \in \mathbb{R}^{n_2\times r}$. 
Furthermore, we need add an extra constraint $\mathbf{X}\mathbf{Q}\mathbf{Q}^\top = \mathbf{X}$ to avoid the trivial solution when $r<n_2$. 
If not, $\mathbf{Q}$ may converge to the singular spaces which are corresponding to smaller singular values. For example, when $r = n_1 < n_2$ and $\mathbf{Q}\in\mathbb{R}^{n_2\times (n_2-r)}$, the optimal solution set of $\mathbf{Q}^*$ for Eq.~(\ref{Lemma_1_equation}) includes the null singular spaces of $\mathbf{X}$, which makes $\mathbf{X}\mathbf{Q} = \mathbf{O}$ hold and the objective function value is 0.
Then we have the following:

\begin{theorem}
	Given a fixed matrix $\mathbf{X}\in\mathbb{R}^{n_1\times n_2}$ with $r=\min\{n_1,n_2 \}$, and its skinny Singular Value Decomposition as $\mathbf{X} = \mathbf{U}\mathbf{\Sigma}\mathbf{V}^\top$ where $\mathbf{U}\in\mathbb{R}^{n_1\times r}$, $\mathbf{\Sigma}\in\mathbb{R}^{r\times r}$, and $\mathbf{V}\in\mathbb{R}^{n_2\times r}$. Then {the matrix of right singular vectors} $\mathbf{V}$ optimizes the following:
	\begin{equation}\label{add_THM_EQ}
	\min_{\mathbf{Q}\in\mathbb{R}^{n_2 \times r}}\|\mathbf{X}\mathbf{Q}\|_{2,1},\quad \text{s.t.}\  \mathbf{Q}^\top\mathbf{Q} = \mathbf{I}, \  \mathbf{X}\mathbf{Q}\mathbf{Q}^\top = \mathbf{X}.
	\end{equation}
	\label{Theorem_From_Lemma_2}
\end{theorem}

The proofs of the above {please see} Appendix~\ref{APP_3}. Theorem~\ref{Theorem_From_Lemma_2} shows that, to minimize the $\ell_{2,1}$ norm $\|\mathbf{X}_{(3)}\mathbf{Q}\|_{2,1}$ w.r.t. $\mathbf{Q}$, we can choose $\mathbf{Q}$ as {the matrix of right singular vectors} of $\mathbf{X}_{(3)}$.

\subsubsection{Details of How To Make $\mathbf{Q}$ Data Dependent}

Through the analyses in Sec.~\ref{Sec:From_VM_to_SM}, we make the selection of $\mathbf{Q}$ data-dependent, and the following definitions shows the details.
\begin{definition}\textbf{(VMTQN: Variance Maximization Tensor Q-Nuclear norm)}\label{def_of_vmtqn}
	Let $\mathcal{X}\in\mathbb{R}^{n_1\times n_2\times n_3}$ be a {third-order tensor} and $\mathbf{Q}$ be an orthogonal matrix. If $\mathcal{G}=\mathcal{X}\times_3 \mathbf{Q}$ and $\mathbf{G}^{(i)}$ denotes the frontal slices of $\mathcal{G}$, then the Variance Maximization Tensor Q-Nuclear norm~(VMTQN) is defined as follows:
	\begin{equation}
		\left\| \mathcal{X} \right\|_{Q,*},\ \text{where } \mathbf{Q}=\argmax_{\mathbf{Q}^\top \mathbf{Q} = \mathbf{I}} \ \operatorname{Variance}\left\{ \left\|\mathbf{G}^{(i)}\right\|_F \right\}.
	\end{equation}
\end{definition}
Note that $\mathbf{Q}$ is determined by $\mathcal{X}$. With the help of Lemma~\ref{Intro_lemma_1}, Lemma~\ref{Intro_lemma_2}, and Theorem~\ref{Theorem_From_Lemma_2}, we can incorporate VMTQN into the low rank recovery model.

\begin{definition}\textbf{(Low Tensor Q-rank model with adaptive $\mathbf{Q}$)} 
	By setting the adaptive $\mathbf{Q}$ module as a low-level sub-problem, the low tensor Q-rank model~(\ref{low_tensor_q_rank_model_with_rank}) is transformed into the following:
	\begin{equation}\label{bilevel_low_rank_model_rank}
	\begin{aligned}
	\min_{\mathcal{X}, \mathbf{Q}}\quad \operatorname{rank}_Q (\mathcal{X}),
	\ s.t.\ \Psi(\mathcal{X}) = \mathcal{Y},\ \mathbf{Q}\in \argmin_{\mathbf{Q}^\top\mathbf{Q} = \mathbf{I}} \|\mathbf{X}_{(3)}\mathbf{Q}\|_{2,1},\ \mathbf{X}\mathbf{Q}\mathbf{Q}^\top = \mathbf{X}.
	\end{aligned}
	\end{equation}
	And the corresponding surrogate model~(\ref{low_tensor_q_rank_model_with_norm}) is also replaced by the following:
	\begin{equation}\label{bilevel_low_rank_model_nuclear}
	\begin{aligned}
	\min_{\mathcal{X}, \mathbf{Q}} \quad \left\| \mathcal{X} \right\|_{Q,*},
	\ s.t.\ \Psi(\mathcal{X}) = \mathcal{Y},\ \mathbf{Q}\in \argmin_{\mathbf{Q}^\top\mathbf{Q} = \mathbf{I}} \|\mathbf{X}_{(3)}\mathbf{Q}\|_{2,1},\ \mathbf{X}\mathbf{Q}\mathbf{Q}^\top = \mathbf{X}.
	\end{aligned}
	\end{equation}
	In Eqs.~(\ref{bilevel_low_rank_model_rank}) and~(\ref{bilevel_low_rank_model_nuclear}), $\mathbf{X}_{(3)}\in\mathbb{R}^{n_1n_2\times n_3}$ denotes the mode-3 unfolding matrix of tensor $\mathcal{X}\in\mathbb{R}^{n_1\times n_2\times n_3}$, and $\mathbf{Q}\in\mathbb{R}^{n_3\times r}$ with $r = \min\{n_1n_2,n_3 \}$. 
	\label{Def_1}
\end{definition}

\begin{definition}
	In fact, Theorem~\ref{Theorem_From_Lemma_2} implies $\mathbf{Q} = \mathbf{V}$, where $\mathbf{V}$ is {the matrix of right singular vectors} of $\mathbf{X}_{(3)}$.
	If we let $\operatorname{PCA}(\mathcal{X},3,r) := \argmin_{\mathbf{Q}^\top\mathbf{Q} = \mathbf{I}_r} \|\mathbf{X}_{(3)}\mathbf{Q}\|_{2,1}$ be the operator to obtain {the matrix of right singular vectors} $\mathbf{Q}\in\mathbb{R}^{n_3\times r}$, where $r = \min\{n_1 n_2, n_3\}$, then the models~(\ref{bilevel_low_rank_model_rank}) and~(\ref{bilevel_low_rank_model_nuclear}) can be abbreviated as follows:
	\begin{eqnarray}
	\min_\mathcal{X}\ \operatorname{rank}_{Q} (\mathcal{X}),\quad s.t.\ \Psi(\mathcal{X}) = \mathcal{Y},\ \mathbf{Q} = \operatorname{PCA}(\mathcal{X},3,r),\label{MODEL_1}\\
	\min_\mathcal{X}\ \left\| \mathcal{X} \right\|_{Q,*},\quad s.t.\ \Psi(\mathcal{X}) = \mathcal{Y},\ \mathbf{Q} = \operatorname{PCA}(\mathcal{X},3,r).\label{MODEL_2}
	\end{eqnarray}
\end{definition}

\begin{remark}
	Notice that $\mathbf{Q} \in \mathbb{R}^{n_3\times r}$ in Eqs.~(\ref{bilevel_low_rank_model_rank}) and~(\ref{bilevel_low_rank_model_nuclear}) may not have full columns, i.e., $r < n_3$. The corresponding definitions of $\operatorname{rank}_Q(\mathcal{X})$ in Eq.~(\ref{tensor_q_rank_cal_tucker_product}) and $\left\| \mathcal{X} \right\|_{Q,*}$ in Eq.~(\ref{q_nuclear}) also change to the sum of $r$ frontal slices instead of $n_3$. Then Theorem~\ref{non_square_theorem} guarantee the validity of Eq~(\ref{bilevel_low_rank_model_nuclear}).
\end{remark}

\begin{remark}\label{remark_select_q}
	In fact, from Appendix~\ref{APP_3} we can see that, $r$ can be chosen as any value that satisfies the condition $\operatorname{rank}(\mathbf{X}_{(3)})\leq r\leq \min\{n_1 n_2, n_3\}$, as long as $\mathbf{Q}\in\mathbb{R}^{n_3\times r}$ contains the whole column space of {the matrix of right singular vectors} $\mathbf{V}$ and is pseudo-invertible to make $\mathcal{X}=\mathcal{X}\times_3 \mathbf{Q}\times_3 \mathbf{Q}^{+}$ hold. 
	\label{Remark_1}
\end{remark}
Within this framework, the orthogonal matrix $\mathbf{Q}$ is related to tensor $\mathcal{X}$. As we analyzed, choosing $\mathbf{Q}$ as {the matrix of right singular vectors} may make $\operatorname{rank}_Q(\mathcal{X})$ as low as possible. In other words, there should be more ``small'' frontal slices of $\mathcal{X}\times_3 \mathbf{Q}$, whose Frobenius norms are close to $0$ to guarantee the low tensor Q-rank structure of data with high probability.

Now the question is whether the function $\| \mathcal{X} \|_{Q,*}$ in Eq.~(\ref{MODEL_2}) is still an envelope of the rank function $\text{rank}_{Q} (\mathcal{X})$ in Eq.~(\ref{MODEL_1}) within an appropriate region. The following theorem shows that even if $\| \mathcal{X} \|_{Q,*}$ is no longer a convex function in the bilevel framework~(\ref{MODEL_2}) since $\mathbf{Q}$ is dependent on $\mathcal{X}$, we can still use it as a surrogate for a lower bound of $\text{rank}_{Q} (\mathcal{X})$ in Eq.~(\ref{MODEL_1}).

\begin{theorem}\label{thm_1}
	{Given a column orthonormal matrix $\mathbf{Q}\in\mathbb{R}^{n_3\times r}$, $r = \min\{n_1 n_2, n_3\}$,} we use $\operatorname{rank}_{PCA}(\mathcal{X})$, $\|\mathcal{X}\|_{PCA,\sigma}$, and $\|\mathcal{X}\|_{PCA,*}$ to abbreviate the corresponding concepts as follows: 
	\begin{eqnarray}
	\operatorname{rank}_{PCA}(\mathcal{X}) := \operatorname{rank}_{Q}(\mathcal{X}),  \text{where } \mathbf{Q} = \operatorname{PCA}(\mathcal{X},3,r),\label{PCA_rank}\\
	\|\mathcal{X}\|_{PCA,\sigma}  := \|\mathcal{X}\|_{Q,\sigma}, \quad  \text{where } \mathbf{Q} = \operatorname{PCA}(\mathcal{X},3,r), \label{PCA_sigma}\\
	\|\mathcal{X}\|_{PCA,*}  := \|\mathcal{X}\|_{Q,*}, \quad \text{where } \mathbf{Q} = \operatorname{PCA}(\mathcal{X},3,r)\label{PCA_nuclear}.
	\end{eqnarray}
	Then within the region of $\mathcal{D} = \{\mathcal{X}\ |\  \|\mathcal{X}\|_{PCA,\sigma} \leq 1\}$, the inequality $\|\mathcal{X}\|_{PCA,*} \leq \operatorname{rank}_{PCA}(\mathcal{X})$ holds. 
	Moreover, for every fixed $\mathbf{Q}$, let $\mathcal{S}_Q$ denote the space $\{\mathcal{X}\ |\  \mathbf{Q} \in \operatorname{PCA}(\mathcal{X},3,r)\}$. Then Theorem~\ref{non_square_theorem} indicates that $\|\mathcal{X}\|_{PCA,*}$ is still the tightest convex envelope of $\operatorname{rank}_{PCA}(\mathcal{X})$ in $\mathcal{S}_Q \cap\mathcal{D}$.
\end{theorem}

\begin{remark}
	For any column orthonormal matrix $\mathbf{Q}\in\mathbb{R}^{n_3\times r}$, 
	the corresponding conclusion also holds as long as $\mathcal{X}\times_3 (\mathbf{Q}\mathbf{Q}^\top) = \mathcal{X}$. That is to say, $\|\mathcal{X}\|_{Q,*} \leq \text{rank}_{Q}(\mathcal{X})$ holds within the region $\{\mathcal{X}\ |\  \|\mathcal{X}\|_{Q,\sigma} \leq 1\}$. \label{Remark_2}
\end{remark}

Theorem~\ref{thm_1} shows that though $\|\mathcal{X}\|_{PCA,*}$ could be non-convex, its function value is always below $\text{rank}_{PCA}(\mathcal{X})$. 
Therefore, model~(\ref{MODEL_2}) can be regarded as a reasonable low rank tensor recovery model. Notice that it is actually a bilevel optimization problem.

\subsection{\textbf{Way II~(MOTQN):} Estimate $\mathbf{Q}$ by Manifold Optimization}\label{Sec:MOTQN}

Recalling the data-dependent low rank recovery model Eq.~(\ref{bilevel_low_tensor_q_rank_model}) with $\mathcal{X}\in\mathbb{R}^{n_1\times n_2\times n_3}$, our main idea is to find a learnable $\mathbf{Q}\in\mathbb{R}^{n_3\times n_3}$ to minimize $\text{rank}_Q(\mathcal{X})$. Inspired by Remark~\ref{Remark_2}, if we let $\mathbf{Q} = \argmin_{\mathbf{Q}^\top \mathbf{Q} = \mathbf{I}} \|\mathcal{X}\|_{Q,*}$ to minimize the surrogate function directly, then we can get the following bilevel model:
\begin{equation}\label{MOTQN_model}
\min_{\mathcal{X},\mathbf{Q}}\ \left\| \mathcal{X} \right\|_{Q,*},\quad s.t.\ \Psi(\mathcal{X}) = \mathcal{Y},\  \mathbf{Q} = \argmin_{\mathbf{Q}^\top \mathbf{Q} = \mathbf{I}} \|\mathcal{X}\|_{Q,*}.
\end{equation}
In Eq.~(\ref{MOTQN_model}), the lower-level problem w.r.t. $\mathbf{Q}$ is actually a Stiefel manifold optimization problem.
Similarly, we can define the corresponding nuclear norm as follows: 
\begin{definition}\textbf{(MOTQN: Manifold Optimization Tensor Q-Nuclear norm)}\label{def_of_motqn}
	Let $\mathcal{X}\in\mathbb{R}^{n_1\times n_2\times n_3}$ be a {third-order tensor} and $\mathbf{Q}\in\mathbb{R}^{n_3\times n_3}$ be an orthogonal matrix. Then the Manifold Optimization Tensor Q-Nuclear norm~(MOTQN) is defined as:
	\begin{equation}\label{DEF_of_MOTQN}
	\left\| \mathcal{X} \right\|_{Q,*},\ \text{where } \mathbf{Q} = \argmin_{\mathbf{Q}^\top \mathbf{Q} = \mathbf{I}} \|\mathcal{X}\|_{Q,*}.
	\end{equation}
\end{definition}

Different from VMTQN, the learnable $\mathbf{Q}$ in Eq.~(\ref{MOTQN_model}) should be a square matrix, i.e., $\mathbf{Q}\in\mathbb{R}^{n_3\times n_3}$. If not, as mentioned in Sec.~\ref{Sec:From_VM_to_SM}, $\mathbf{Q}$ may converge to the singular spaces which are corresponding to smaller singular values. To avoid this case, we let $\mathbf{Q}\in\mathbb{R}^{n_3\times n_3}$.
Following, the key point of solving this model is how to deal with such an orthogonality constrained optimization problem:
\begin{equation}\label{low_level_problem}
	\mathbf{Q} = \argmin_{\mathbf{Q}^\top \mathbf{Q} = \mathbf{I}} \|\mathcal{X}\|_{Q,*} = \argmin_{\mathbf{Q}^\top \mathbf{Q} = \mathbf{I}} \sum_{i=1}^{n_3} \left\| \mathbf{G}^{(i)} \right\|_*,\quad \text{where } \mathcal{G} = \mathcal{X} \times_3 \mathbf{Q}.
\end{equation}
Note that Eq.~(\ref{low_level_problem}) is actually a non-convex problem due to the orthogonality constraint.
The usual way is to perform the manifold Gradient Descent on the Stiefel manifold, which evolves along the manifold geodesics~\cite{edelman1998geometry}. However, this method usually requires a lot of computation to calculate the projected gradient direction of the objective function. Meanwhile, the work~\cite{wen2013feasible} develops a technique to solve such orthogonality constrained problem \textbf{iteratively}, which generates feasible points by the Cayley transformation and only involves {matrix multiplication and inversion.} 
Here we consider to use their algorithm to solve the low-level problem.

\subsubsection{Optimization with Orthogonality Constraints}

Assume $\mathbf{Q}\in\mathbb{R}^{n\times r}$ and {denote the gradient of the objective function $f(\mathbf{Q}) = \|\mathcal{X}\|_{Q,*}$ w.r.t. $\mathbf{Q}$ at $\mathbf{Q}_k$~(the $k$-th iteration) by $\mathbf{P}\in\mathbb{R}^{n\times r}$}. Then the projection of $\mathbf{P}$ in the tangent space of the Stiefel manifold at $\mathbf{Q}_k$ is $\mathbf{A}\mathbf{Q}_k$, where $\mathbf{A} = \mathbf{P}\mathbf{Q}_k^\top - \mathbf{Q}_k\mathbf{P}^\top$ and $\mathbf{A}\in\mathbb{R}^{n\times n}$~\cite{wen2013feasible}. Instead of parameterizing the geodesic of the Stiefel manifold along direction $\mathbf{A}$ using the exponential function, inspired by~\cite{wen2013feasible}, we generate feasible points by the following Cayley transformation:
\begin{equation}\label{CT_equation}
	\mathbf{Q}(\tau) = \mathbf{C}(\tau) \mathbf{Q}_k, \qquad \text{where}\quad \mathbf{C}(\tau) = \left( \mathbf{I} + \frac{\tau}{2} \mathbf{A}\right)^{-1} \left(\mathbf{I} - \frac{\tau}{2} \mathbf{A} \right),
\end{equation}
where $\mathbf{I}$ is the identity matrix and $\tau \in \mathbb{R}$ is a parameter to determine the step size of $\mathbf{Q}_{k+1}$. That is to say, $\mathbf{Q}(\tau)$ is a re-parameterized geodesic w.r.t. $\tau$ on the Stiefel manifold. 
Moreover, if $\mathbf{Q}_k^\top\mathbf{Q}_k=\mathbf{I}$ holds, then $\mathbf{Q}(\tau)$ has the following properties:
\newline
\newline
\begin{enumerate*}[label=(\arabic*),itemjoin={\ }]
		\item $\frac{d}{d\tau}\mathbf{Q}(0) = -\mathbf{A}\mathbf{Q}_k$,
		\item $\mathbf{Q}(\tau)$ is smooth in $\tau$,
		\item $\mathbf{Q}(0)= \mathbf{Q}_k$,
		\item $\mathbf{Q}(\tau)^\top\mathbf{Q}(\tau) = \mathbf{I}$.
\end{enumerate*}
\newline
\newline
The work~\cite{wen2013feasible} shows that if $\tau$ is in a proper range, $\mathbf{Q}(\tau)$ can lead to a lower objective function value than $\mathbf{Q}(0)$ on the Stiefel manifold. 
In summery, solving the problem $\mathbf{Q} = \argmin_{\mathbf{Q}^\top \mathbf{Q} = \mathbf{I}} \|\mathcal{X}\|_{Q,*}$ consists of two steps: (1) find a proper $\tau^*$ to make the value of the objective function $f(\mathbf{Q}(\tau)) = \|\mathcal{X}\|_{Q(\tau),*}$ decrease; (2) update $\mathbf{Q}_{k+1}$ by Eq.~(\ref{CT_equation}), i.e., $\mathbf{Q}_{k+1} = \mathbf{Q}(\tau^*)$.

\subsubsection{Details of How To Estimate $\tau^*$ and Update $\mathbf{Q}_k$}

\textbf{(1):} We first compute the gradient of the objective function $f(\mathbf{Q}) = \|\mathcal{X}\|_{Q,*}$ w.r.t. $\mathbf{Q}$ at $\mathbf{Q}_k$. According to the chain rule, we get the following:
\begin{equation}\label{partial_f_to_Q}
\frac{\partial f(\mathbf{Q})}{\partial \mathbf{Q}} = \frac{\partial \mathcal{G}}{\partial \mathbf{Q}} \cdot \frac{\partial f(\mathbf{Q})}{\partial \mathcal{G}} = \frac{\partial( \mathbf{G_{(3)}})}{\partial\mathbf{Q}}\times \text{unfold}_3\left(\frac{\partial f(\mathbf{Q})}{\partial \mathcal{G}}\right).
\end{equation}

Note that $\mathcal{G} = \mathcal{X}\times_3 \mathbf{Q}$ and $\mathbf{G}_{(3)} = \mathbf{X}_{(3)} \mathbf{Q}$, then we can get $\frac{\partial \mathbf{G_{(3)}}}{\partial \mathbf{Q}} = \mathbf{X}_{(3)}^\top$ where $\mathbf{G}_{(3)}$ and $\mathbf{X}_{(3)}$ are the mode-3 unfolding matrices.
Additionally, Eq.~(\ref{low_level_problem}) shows that $f(\mathbf{Q}) = \sum_{i=1}^{n_3} \| \mathbf{G}^{(i)} \|_*$ where $\mathbf{G}^{(i)} $ are the frontal slices of $\mathcal{G}$. 
We let $\mathbf{H}^{(i)} = \mathbf{U}^{(i)} \mathbf{V}^{(i)}$, where $\mathbf{H}^{(i)}$ denotes the frontal slice of $\mathcal{H}$ and $\mathbf{U}^{(i)}\mathbf{V}^{(i)}$ denotes the left/right singular matrices of $\mathbf{G}^{(i)}$ by skinny SVD~\cite{petersen2008matrix}.
Therefore, $\mathcal{H} = \frac{\partial f(\mathbf{Q})}{\partial \mathcal{G}}$ {is the same as the matrix case and can be obtained from the singular value decomposition}\footnote{The subgradient of matrix nuclear norm $\|\mathbf{M}\|_*$ w.r.t. $\mathbf{M}$ is $\{ \mathbf{U}\mathbf{V}^\top + \mathbf{W}\ \|\ \mathbf{U}^\top \mathbf{W} = \mathbf{O}, \mathbf{W}\mathbf{V} = \mathbf{O}, \|\mathbf{W}\|\leq 1 \}$, where $\mathbf{M} = \mathbf{U}\Sigma\mathbf{V}^\top$ is the SVD of $\mathbf{M}$.}. 

In summary, the gradient of the objective function $f(\mathbf{Q})$ w.r.t. $\mathbf{Q}$ at $\mathbf{Q}_k$~(denoted by $\mathbf{P}$) can be written as follows:
\begin{equation}\label{gradient}
	\text{Gradient}= \mathbf{P} = \frac{\partial f(\mathbf{Q})}{\partial \mathbf{Q}} = \frac{\partial \mathcal{G}}{\partial \mathbf{Q}}\cdot \frac{\partial f(\mathbf{Q})}{\partial \mathcal{G}} = \mathbf{X}_{(3)}^\top \mathbf{H}_{(3)}.
\end{equation}
where $\mathbf{X}_{(3)}$ and $\mathbf{H}_{(3)}$ are the mode-$3$ unfolding matrices of $\mathcal{X}$ and $\mathcal{H}$, respectively.
\newline
\textbf{(2):} Then we construct a geodesic curve along the gradient direction on the Stiefel manifold by Eq.~(\ref{CT_equation}):
\begin{equation}\label{construct_curve}
\mathbf{Q}(\tau) = \left( \mathbf{I} + \frac{\tau}{2} \mathbf{A}\right)^{-1} \left(\mathbf{I} - \frac{\tau}{2} \mathbf{A} \right) \mathbf{Q}_k, \quad \text{where}\quad \mathbf{A} = \mathbf{P}\mathbf{Q}_k^\top - \mathbf{Q}_k\mathbf{P}^\top.
\end{equation}
We consider the following problem for finding a proper $\tau$:
\begin{equation}
	\tau^* = \argmin_{0\leq \tau \leq \varepsilon} f(\mathbf{Q}(\tau)) = \argmin_{0\leq \tau \leq \varepsilon} g(\tau) = \argmin_{0\leq \tau \leq \varepsilon} \|\mathcal{X}\|_{Q(\tau),*},
\end{equation}
where $\varepsilon$ is a given parameter to ensure that $\tau^*$ is small enough and  $\|\frac{\tau}{2}\mathbf{A}\|\leq 1$ holds. Then we can simplify $g(\tau) = f(\mathbf{Q}(\tau))$ with the equation $\left( \mathbf{I} + \frac{\tau}{2} \mathbf{A}\right)^{-1} = \mathbf{I} + \sum_{l=1}^{\infty}\left( -\frac{\tau}{2}\mathbf{A} \right)^l$ and obtain the following:
\begin{equation}
	g(\tau) = f(\mathbf{Q}(\tau)) = f\left(\left( \mathbf{I} + 2\sum_{l=1}^{\infty}\left( -\frac{\tau}{2}\mathbf{A} \right)^l \right)\mathbf{Q}_k\right) \approx f\left( \left( \mathbf{I} - \tau \mathbf{A}+ \frac{\tau^2}{2}\mathbf{A}^2 \right)\mathbf{Q}_k \right).
\end{equation}

Given that $\tau^*$ is small enough, we can approximate $g(\tau)$ via its second order Taylor expansion at $\tau=0$, i.e., $g(\tau) = g(0) + g'(0)\cdot\tau + \frac{1}{2} g''(0)\cdot\tau^2 + \mathcal{O}(\tau^3)$. It should be mentioned that since $f(\mathbf{Q})$ is non-convex w.r.t. $\mathbf{Q}$, the sign of $g''(0)$ is uncertain. However, Wen et al~\cite{wen2013feasible} point out that $g'(0)=-\frac{1}{2}\|\mathbf{A}\|_F^2$ always holds. Thus we can estimate an optimal solution $\tau^*$ via:
\begin{equation}\label{determine_tau_star}
	\tau^* = \min\{\varepsilon, \tilde{\tau}\},\quad \text{where } \varepsilon < \frac{2}{\|\mathbf{A}\|}, \text{and } \tilde{\tau} = 
	\begin{cases}
	-\frac{g'(0)}{g''(0)}, &g''(0) > 0\\
	\frac{1}{\|\mathbf{A}\|}, &g''(0) \leq 0.
	\end{cases}\\
\end{equation}
Here we give the following Lemma to omit the calculation process~(See Appendix~\ref{APP_4}).

\begin{lemma}\label{first_and_second_d_of_f}
	Let $g(\tau) = f(\mathbf{Q}(\tau)) = \|\mathcal{X}\|_{Q(\tau),*}$ and $\mathbf{Q}(\tau) \approx \left( \mathbf{I} - \tau \mathbf{A}+ \frac{\tau^2}{2}\mathbf{A}^2 \right)\mathbf{Q}_k $, where $\mathbf{A}$ is defined in Eq.~(\ref{construct_curve}).
	Then the first and the second order derivatives of $g(\tau)$ evaluated at 0 can be estimated as follows:
	\begin{equation}
		\begin{aligned}
		g'(0)  \approx  \left\langle \mathbf{X}_{(3)}^\top \mathbf{H}_{(3)}, -\mathbf{A} \mathbf{Q}_k \right\rangle, \text{ and }
		g''(0) \approx  \left\langle \mathbf{X}_{(3)}^\top \mathbf{H}_{(3)}, \mathbf{A}^2\mathbf{Q}_k \right\rangle, \\
		\end{aligned}
	\end{equation}
	where $\mathbf{X}_{(3)}$ and $\mathbf{H}_{(3)}$ are defined as the same in Eq.~(\ref{gradient}).
\end{lemma}

By using Eq.~(\ref{determine_tau_star}) and Lemma~\ref{first_and_second_d_of_f}, we can obtain the optimal step size $\tau^*$ and then use Eq.~(\ref{construct_curve}) to update $\mathbf{Q}_{k+1} = \mathbf{Q}(\tau^*)$. Algorithm~\ref{CT_to_update_q_k} organizes the whole calculation process.

Back to the bilevel low rank tensor recovery model Eq.~(\ref{MOTQN_model}), for the lower-level problem Eq.~(\ref{low_level_problem}), we finish the iterative updating step by Algorithm~\ref{CT_to_update_q_k}. Once $\mathbf{Q}_{k+1}$ is fixed, the upper-level problem can be solved easily.

\begin{algorithm}[t]
	\caption{\small Updating $\mathbf{Q}$ iteratively to solve Eq.~(\ref{DEF_of_MOTQN}).} 
	\textbf{Input:} Tensor $\mathcal{X} \in \mathbb{R}^{n_1\times n_2\times n_3}$, orthogonal matrix $\mathbf{Q_0}\in\mathbb{R}^{n_3\times n_3}$.
	\begin{enumerate}
		\item \textbf{while not convergence}
		\item  \quad Calculate $\mathbf{P} = \mathbf{X}_{(3)}^\top \mathbf{H}_{(3)}$ by Eq.~(\ref{gradient}).
		\item  \quad Calculate $\mathbf{A} = \mathbf{P}\mathbf{Q}_k^\top - \mathbf{Q}_k\mathbf{P}^\top$ by Eq.~(\ref{construct_curve}).
		\item  \quad Estimate $\tau^* = \min\{\varepsilon, \tilde{\tau}\}$  by Eq.~(\ref{determine_tau_star}) and Lemma~\ref{first_and_second_d_of_f}.
		\item  \quad Update $\mathbf{Q}_{k+1} = \mathbf{Q}(\tau^*)$ by Eq.~(\ref{construct_curve}).
		\item \textbf{end while}
	\end{enumerate}
	\textbf{Output:} Matrix $\mathbf{Q}_K$.
	\label{CT_to_update_q_k} 
\end{algorithm}

\subsection{Applicability of VMTQN and MOTQN}\label{Sec:two_case}

In Sec.\ref{Sec:MOTQN}~(MOTQN), we mention that $\mathbf{Q}\in\mathbb{R}^{n_3\times n_3}$ should be a square matrix but not in Sec.\ref{Sec:VMTQN}~(VMTQN). In this section, we start from this point and analyze the impact of the size of $\mathcal{X}\in\mathbb{R}^{n_1\times n_2\times n_3}$ on the applicability of these two methods.

\subsubsection{Case 1: $r = n_1n_2 \ll n_3$}

In this case, VMTQN model in Eq.~(\ref{MODEL_2}) usually performs better than other methods in terms of computational efficiency, including MOTQN and other works~\cite{excel_20,xu2019fast,song2019robust,lu2019cvpr,jiang2020framelet}. As we can see from Sec.\ref{Sec:VMTQN} of VMTQN model, we need to calculate a skinny right singular matrix $\mathbf{V}$ of an unfolding matrix $\mathbf{X}_{(3)}\in\mathbb{R}^{n_1n_2\times n_3}$. If $r < n_3$, then not only the computational complexity is not too large, but $\mathbf{Q}$ can play the role of feature selection like Principal Component Analysis, which corresponds to the notation $\mathbf{Q} = \operatorname{PCA}(\mathcal{X},3,r)$. 

Meanwhile, MOTQN and the work~\cite{excel_20,xu2019fast,song2019robust,lu2019cvpr} usually need to have a square factor matrix $\mathbf{Q}$, even that~\cite{jiang2020framelet} requires the columns of $\mathbf{Q}$ to be redundant.

\subsubsection{Case 2: $n_1n_2 > n_3 = r$ or even have the same order of magnitude}
\label{Sec:case_2_Applicability}
In this case, MOTQN model in Eq.~(\ref{MOTQN_model}) has the best explainability and rationality. On the one hand, with the same size of $\mathbf{Q}\in\mathbb{R}^{n_3\times n_3}$, MOTQN minimize the tensor Q-nuclear norm directly, which corresponds to the definition of low rank structure properly. On the other hand, thanks to the algorithm in~\cite{wen2013feasible}, the optimization of MOTQN model has good convergence guarantee.

\section{Applications to Tensor Completion}	
\subsection{Low Rank Tensor Completion Model}
In the {third-order tensor} tensor completion task, $\Omega$ is an index set consisting of the indices $\{(i,j,k)\}$ which can be observed, and the operator $\Psi$ in Eqs.~(\ref{MODEL_1}) and~(\ref{MODEL_2}) is replaced by an orthogonal
projection operator $\mathcal{P}_\Omega$, where $\mathcal{P}_\Omega(\mathcal{X}_{ijk}) = \mathcal{X}_{ijk}$ if $(i,j,k)\in\Omega$ and $0$ otherwise. The observed tensor $\mathcal{Y}$ satisfies $\mathcal{Y} = \mathcal{P}_\Omega(\mathcal{Y})$.
Then the tensor completion model based on our two ways are given by:
\begin{equation}\label{tc_bilevel_model}
\textbf{(VMTQN):}\quad  \min_\mathcal{X}\ \left\| \mathcal{X} \right\|_{Q,*}, \quad s.t.\ \mathcal{P}_\Omega(\mathcal{X}) = \mathcal{Y},\ \mathbf{Q} = \operatorname{PCA}(\mathcal{X},3,r),
\end{equation}
and
{
\begin{eqnarray}\label{TC_MOTQN}
	\textbf{(MOTQN):}\quad &  \min_{\mathcal{X},\mathbf{Q}}\ &\left\| \mathcal{X} \right\|_{Q,*} \notag \\
	& s.t.\ &\mathbf{Q} = \argmin_{\mathbf{Q}^\top \mathbf{Q} = \mathbf{I}} \|\mathcal{X}\|_{Q,*},\ \mathcal{P}_\Omega(\mathcal{X}) = \mathcal{Y}.\quad\quad\quad\quad\quad\quad
\end{eqnarray}
}
where $\mathcal{X}$ is the tensor that has low rank structure. In Eq.~(\ref{tc_bilevel_model}), $\mathbf{Q}\in\mathbb{R}^{n_3\times r}$ is an column orthonormal matrix with $r = \min\{n_1 n_2, n_3\}$. While in Eq.~(\ref{TC_MOTQN}), $\mathbf{Q}\in\mathbb{R}^{n_3\times n_3}$ is a square orthogonal matrix.
To solve these models by ADMM based method~\cite{LuADMMPAMI}, we introduce an intermediate tensor $\mathcal{E}$ to separate $\mathcal{X}$ from $\mathcal{P}_\Omega(\cdot)$. Let $\mathcal{E} = \mathcal{P}_\Omega(\mathcal{X}) - \mathcal{X}$, then $\mathcal{P}_\Omega(\mathcal{X}) = \mathcal{Y}$ is translated to $\mathcal{X} + \mathcal{E} = \mathcal{Y},\ \mathcal{P}_{\Omega}(\mathcal{E}) = \mathcal{O}$, where $\mathcal{O}$ is an all-zero tensor. Then we get the following two models:
\begin{equation}\label{Eq:VMTQN_TC}
	\textbf{(VMTQN):}\ \min_{\mathcal{X},\mathcal{E},\mathbf{Q}}\ \left\| \mathcal{X} \right\|_{Q,*},
	\ s.t.\ \mathcal{X} + \mathcal{E} = \mathcal{Y},\  \mathcal{P}_\Omega(\mathcal{E}) = \mathcal{O},\ \mathbf{Q} =	\operatorname{PCA}(\mathcal{X},3,r),
\end{equation}
and
\begin{equation}\label{Eq:MOTQN_TC}
	\textbf{(MOTQN):}\ \min_{\mathcal{X},\mathcal{E},\mathbf{Q}}\ \left\| \mathcal{X} \right\|_{Q,*},
	\ s.t.\ \mathcal{X} + \mathcal{E} = \mathcal{Y},\  \mathcal{P}_\Omega(\mathcal{E}) = \mathcal{O},\ \mathbf{Q} = \argmin_{\mathbf{Q}^\top \mathbf{Q} = \mathbf{I}} \|\mathcal{X}\|_{Q,*}.
\end{equation}
Note that in Eq.~(\ref{Eq:MOTQN_TC}), the constraint $\mathbf{Q} = \argmin_{\mathbf{Q}^\top \mathbf{Q} = \mathbf{I}} \|\mathcal{X}\|_{Q,*}$ is the same as the objective function, thus it can be omitted. Nevertheless, in order to keep Eqs.~(\ref{Eq:VMTQN_TC}) and~(\ref{Eq:MOTQN_TC}) unified in form and express the dependence of $\mathbf{Q}$ and $\mathcal{X}$ conveniently, we reserve this constraint here.

\subsection{Optimization Algorithm}
Since $\mathbf{Q}$ is dependent on $\mathcal{X}$, it is difficult to solve the models~(\ref{Eq:VMTQN_TC}) and~(\ref{Eq:MOTQN_TC}) w.r.t. $\{\mathcal{X},\mathbf{Q}\}$ directly.
Here we adopt the idea of alternating minimization to solve $\mathcal{X}$ and $\mathbf{Q}$ alternately. We separate the sub-problem of solving $\mathbf{Q}$ as a sub-step in every $K$-iteration, and then update $\mathcal{X}$ with a fixed $\mathbf{Q}$ by the ADMM method~\cite{LuADMMPAMI,LuIJCAI2018}. 
The partial augmented Lagrangian function of Eqs.~(\ref{Eq:VMTQN_TC}) and~(\ref{Eq:MOTQN_TC}) is
\begin{equation}\label{partial_augmented_Lagrangian}
L(\mathcal{X},\mathcal{E},\mathcal{Z},\mu) = \left\| \mathcal{X} \right\|_{Q,*} + \left\langle \mathcal{Z}, \mathcal{Y} - \mathcal{X} - \mathcal{E} \right\rangle + \frac{\mu}{2} \left\| \mathcal{Y} - \mathcal{X} - \mathcal{E} \right\|_F^2,
\end{equation}
where $\mathcal{Z}$ is the dual variable and $\mu > 0$ is the penalty parameter. Then we can update each component $\mathbf{Q}$, $\mathcal{X}$, $\mathcal{E}$, and $\mathcal{Z}$ alternately. Algorithms~\ref{Algorithm_1} and~\ref{Algorithm_2} show the details about the optimization methods to Eqs.~(\ref{Eq:VMTQN_TC}) and~(\ref{Eq:MOTQN_TC}). 
In order to improve the efficiency and stable convergence of the algorithm, we introduce a parameter $K$ to control the update frequency of $\mathbf{Q}$ with the help of heuristic design. The different effects of $K$ on the two models are explained in Sec.~\ref{Sec:Convergence} and Sec.~\ref{Sec:Complexity}, respectively.

Note that there is one operator $\mathbf{Prox}$ in the sub-step of updating $\mathcal{X}$ as follows:
\begin{equation}\label{prox_ope_q}
\mathcal{X} = \mathbf{Prox}_{\lambda, \|\cdot\|_{Q,*}}\left(\mathcal{T}\right) := \argmin_{\mathcal{X}} \lambda\|\mathcal{X}\|_{Q,*} + \frac{1}{2} \left\| \mathcal{X} - \mathcal{T}  \right\|_F^2,
\end{equation}
where $\mathbf{Q}\in\mathbb{R}^{n_3\times r}$ is a given column orthonormal matrix and $\|\mathcal{X}\|_{Q,*}$ is the tensor Q-nuclear norm of $\mathcal{X}$ which is defined in Eq.~(\ref{q_nuclear}). Algorithm~\ref{Algorithm_2} shows the details of solving this operator.

\begin{algorithm}[t]
	\caption{\small Solving the problems~(\ref{Eq:VMTQN_TC}) and~(\ref{Eq:MOTQN_TC}):VMTQN and MOTQN models by ADMM.} 
	\textbf{Input:} Observation samples $\mathcal{Y}_{ijk}$, $(i,j,k)\in\Omega$, of tensor $\mathcal{Y}\in\mathbb{R}^{n_1\times n_2\times n_3}$.\\ 
	\textbf{Initialize:}  $\mathcal{X}_0,\  \mathcal{E}_0,\ \mathcal{Z}_0,\ \mathbf{Q}_0\in\mathbb{R}^{n_3\times r}$. Parameters $\ k = 1,\ \rho > 1,\ \mu_{0},\ \mu_{max},\ \varepsilon,\ K$.\\ 
	\textbf{While} not converge \textbf{do} 
	\begin{enumerate} 
		\item Update $\mathbf{Q}_k$~ by one of the following:
		\begin{equation}\label{Update_Q}
		\textbf{(VMTQN):}\quad\mathbf{Q}_k =
		\begin{cases}
		\mathbf{Q}_{k-1},\quad\quad\quad & k \text{ mod } K \neq K-1,\\
		\text{PCA}\left(\mathcal{Y}-\mathcal{E}_{k-1}+\frac{\mathcal{Z}_{k-1}}{\mu_{k-1}},3, r\right), &k \text{ mod } K = K-1 .
		\end{cases}
		\end{equation}
		\begin{equation}\label{Update_Q_MO}
		\textbf{(MOTQN):}\quad\mathbf{Q}_k =
		\begin{cases}
		\mathbf{Q}_{k-1},\quad\quad\quad & k \text{ mod } K \neq K-1,\\
		\mathbf{Q}(\tau^*)\text{\quad by using Algorithm~\ref{CT_to_update_q_k}}, &k \text{ mod } K = K-1 .
		\end{cases}
		\end{equation}
		\item Update $\mathcal{X}_k$ by
		\begin{equation}\label{Update_X}
		\mathcal{X}_k = \mathbf{Prox}_{\mu_{k-1}^{-1}, \|\cdot\|_{Q_k,*}}\left(\mathcal{Y}-\mathcal{E}_{k-1}+\frac{\mathcal{Z}_{k-1}}{\mu_{k-1}}\right).
		\end{equation}
		\item Update $\mathcal{E}_k$ by 
		\begin{equation}
		\mathcal{E}_k = \mathcal{P}_{\Omega^\complement}\left( \mathcal{Y} - \mathcal{X}_k + \frac{\mathcal{Z}_{k-1}}{\mu_{k-1}}\right),\quad 
		\end{equation}
		$\text{where } \Omega^\complement \text{ is the complement of } \Omega.$
		\item Update the dual variable $\mathcal{Z}_k$ by \begin{equation}
		\mathcal{Z}_k = \mathcal{Z}_{k-1} + \mu_{k-1}\left( \mathcal{Y} - \mathcal{X}_k - \mathcal{E}_k \right).
		\end{equation}
		\item Update $\mu_k$ by
		\begin{equation}\label{Update_mu}
		\mu_k = \min\{\rho\mu_{k-1},\mu_{max}\}.
		\end{equation}
		\item Check the convergence condition: $\left\| \mathcal{X}_k - \mathcal{X}_{k-1} \right\|_{\infty} \leq \varepsilon$, $\left\| \mathcal{E}_k - \mathcal{E}_{k-1} \right\|_{\infty} \leq \varepsilon$, and $\left\| \mathcal{Y} - \mathcal{X}_{k} - \mathcal{E}_k \right\|_{\infty} \leq \varepsilon$.
		\item $k \leftarrow k+1$.
	\end{enumerate} 
	\textbf{end While}\\ 	
	\textbf{Output:} The target tensor $\mathcal{X}_k$.
	\label{Algorithm_1} 
\end{algorithm}

\begin{algorithm}[t]
	\caption{\small Solving the proximal operator $\mathbf{Prox}_{\lambda, \|\cdot\|_{Q,*}}\left(\mathcal{T}\right)$ in Eq.~(\ref{prox_ope_q}) and~(\ref{Update_X}).} 
	\textbf{Input:} Tensor $\mathcal{T} \in \mathbb{R}^{n_1\times n_2\times n_3}$, column orthonormal matrix $\mathbf{Q}\in\mathbb{R}^{n_3\times r}$.
	\begin{enumerate}
		\item  $\mathcal{G} = \mathcal{T} \times_3 \mathbf{Q}$.
		\item  \textbf{for} $i = 1 \text{ to } r$:
		
		\quad$[\mathbf{U},\mathbf{S},\mathbf{V}] = \operatorname{SVD}(\mathbf{G}^{(i)})$.
		
		\quad$\mathbf{G}^{(i)} = \mathbf{U}(\mathbf{S} - \lambda \mathbf{I})_+ \mathbf{V}^\top$,\quad where $(x)_+ = \max\{x,0\}$.
		\item \textbf{end for}
		\item $\mathcal{X} = \mathcal{G}\times_3 \mathbf{Q}^\top + \mathcal{T}\times_3 (\mathbf{I} - \mathbf{Q}\mathbf{Q}^\top).$
	\end{enumerate}
	
	\textbf{Output:} Tensor $\mathcal{X}$.
	\label{Algorithm_2} 
\end{algorithm}

\subsection{Convergence Analysis}\label{Sec:Convergence}

\subsubsection{VMTQN Model}
For the models~(\ref{tc_bilevel_model}) or~(\ref{Eq:VMTQN_TC}), it is hard to analyze the convergence of the corresponding optimization method directly. The constraint on $\mathbf{Q}$ is non-linear and the objective function is essentially non-convex w.r.t. $\mathbf{Q}$, which increase the difficulty of analysis.
However, the conclusions of~\cite{LuADMMPAMI,excel_26,excel_46,Lin2011NIPS,OPTALG} guarantee the convergence to some extent. 

In practical applications, we can fix $\mathbf{Q}_k=\mathbf{Q}$ in every $K$ iterations to solve a convex problem w.r.t. $\mathcal{X}$. As long as $\mathcal{X}$ is convergent, by using the following Lemma~\ref{partial_lemma}, the change of $\mathbf{Q}$ is bounded. 
\begin{lemma}\label{partial_lemma}~\cite{petersen2008matrix}
	Given a matrix $\mathbf{X}$ and its Singular Value Decomposition $\mathbf{X} = \mathbf{U}\mathbf{\Sigma}\mathbf{V}^\top$. Let $\mathbf{v}_i$ denotes the $i$-th column of matrix $\mathbf{V}$ and $\sigma_j$ denotes the $j$-th singular value of matrix $\mathbf{X}$. {Denote the sub-differential of a variable by $\partial(\cdot)$}, then we have the following:
	\begin{equation}\label{partial_relation}
	\partial(\mathbf{v}_i) = \left( \sigma_i^2 \mathbf{I} - \mathbf{X}^\top \mathbf{X}  \right)^+ \partial(\mathbf{X}^\top \mathbf{X}) \mathbf{v}_i.
	\end{equation}
	If $v_{ij}$ represents the $j$-th element of $\mathbf{v}_i$, then $\left\|\frac{\partial (v_{ij})}{\partial(\mathbf{X}^\top \mathbf{X})}\right\|_2 < \infty$.
\end{lemma}

Lemma~\ref{partial_lemma} indicates that, as long as the change of $\mathcal{X}$ is bounded by penalty term with proper $K$ and $\rho$, the change of $\mathbf{Q}$ will also be bounded to some extent. Then $\lim\limits_{k\rightarrow\infty} Q_k \approx \operatorname{PCA}(\mathcal{X}_k,3,r)$ gradually meets the constraints.

{Unfortunately, Updating the variable $\mathbf{Q}_k$ in Eq.~(\ref{Update_Q}) needs to solve a singular linear system, while the objective norm~$\|\mathcal{X}\|_{Q,*}$ in Eq.~(\ref{Eq:VMTQN_TC}) is non-convex w.r.t. $\mathbf{Q}$. Therefore, it is difficult to prove the conclusion that the Lagrangian function in Eq.~(\ref{partial_augmented_Lagrangian}) of Algorithm~\ref{Algorithm_1} decreases strictly in each iteration.  }
However, we give another Theorem that the iterations corresponding to Eqs.~(\ref{Update_X})-(\ref{Update_mu}) are convergent in the case of fixed $\mathbf{Q}$.

\begin{theorem}\label{Cor_1}
	Given a fixed $\mathbf{Q}$ in every $K$ iterations, the tensor completion model~(\ref{Eq:VMTQN_TC}) can be solved effectively by Algorithm~\ref{Algorithm_1} with $\mathbf{Q}_k = \mathbf{Q}$ in Eq.~(\ref{Update_Q}), where $\Psi$ is replaced by $\mathcal{P}_\Omega$. The rigorous convergence guarantees can be obtained directly due to the convexity as follows:
	
	Let $(\mathcal{X}^*,\mathcal{E}^*,\mathcal{Z}^*)$ be one KKT point of problem~(\ref{Eq:VMTQN_TC}) with fixed $\mathbf{Q}$, $\hat{\mathcal{X}}_K=\frac{\sum_{k=0}^K\frac{1}{\mu_k}\mathcal{X}_{k+1}}{\sum_{k=0}^K\frac{1}{\mu_k}}$, and $\hat{\mathcal{E}}_K=\frac{\sum_{k=0}^K\frac{1}{\mu_k}\mathcal{E}_{k+1}}{\sum_{k=0}^K\frac{1}{\mu_k}}$, then we have
	\begin{equation}
	\|\hat{\mathcal{X}}_{K+1}+\hat{\mathcal{E}}_{K+1}-\mathcal{Y}\|_F^2\leq O\left(\frac{1}{\sum_{k=0}^K\frac{1}{\mu_k}}\right),
	\end{equation}
	and
	\begin{equation}
	0\leq\|\hat{\mathcal{X}}_{K+1}\|_{Q,*}-\|\mathcal{X}^*\|_{Q,*}+ \left\langle \mathcal{Z}^*,\hat{\mathcal{X}}_{K+1}+\hat{\mathcal{E}}_{K+1}-\mathcal{Y}\right\rangle \leq O\left(\frac{1}{\sum_{k=0}^K\frac{1}{\mu_k}}\right).
	\end{equation}
\end{theorem}

\subsubsection{MOTQN Model}

Different from VMTQN model, as we mentioned in Sec.\ref{Sec:case_2_Applicability}, MOTQN model has a complete guarantee of convergence with the help of~\cite{wen2013feasible}. The updating step in Eq.~(\ref{Update_Q_MO}) can strictly guarantee the decrease of the objective function value $\left\| \mathcal{X} \right\|_{Q,*}$ with a proper step size $\tau^*$.
\begin{lemma}\label{lemma_of_CT}(Lemma~$3$ of~\cite{wen2013feasible})
	{Denote the gradient of the objective function $f(\mathbf{Q})$ w.r.t. $\mathbf{Q}$ at $\mathbf{Q}_k$ by $\mathbf{P}$} and let $\mathbf{A} = \mathbf{P}\mathbf{Q}_k^\top - \mathbf{Q}_k\mathbf{P}^\top$ be a skew-symmetric matrix. If we define $\mathbf{Q}(\tau)$ by Eq.~(\ref{construct_curve}), then $\mathbf{Q}(\tau)$ is a descent curve at $\tau=0$, that is,
	\begin{equation}
		f'_\tau(\mathbf{Q}(0)) := \frac{\partial f(\mathbf{Q}(\tau))}{\partial \tau} \Big|_{\tau = 0} = -\frac{1}{2} \left\|\mathbf{A}\right\|_F^2 \leq 0.
	\end{equation}
\end{lemma}

Lemma~\ref{lemma_of_CT} indicates that, as long as $\tau$ is small enough,  Eq.~(\ref{Update_Q_MO}) usually decreases the value of $f(\mathbf{Q}(\tau))$. Notice that Eq.~(\ref{partial_augmented_Lagrangian}) is a partial augmented Lagrangian function, hence the value of Lagrangian function will also decreases after Eq.~(\ref{Update_Q_MO}). Therefore, we have the following theorem to ensure the convergence of Algorithm~\ref{Algorithm_1}:
\begin{theorem}
	{Denote the augmented Lagrangian function of low rank tensor recovery model~\ref{TC_MOTQN} by $L(\mathbf{Q},\mathcal{X},\mathcal{E},\mathcal{Z},\mu) $, } which is shown as follows:
	\begin{equation}\label{THM_5_ALF}
		L(\mathbf{Q},\mathcal{X},\mathcal{E},\mathcal{Z},\mu) = \left\| \mathcal{X} \right\|_{Q,*} + \left\langle \mathcal{Z}, \mathcal{Y} - \mathcal{X} - \mathcal{E} \right\rangle + \frac{\mu}{2} \left\| \mathcal{Y} - \mathcal{X} - \mathcal{E} \right\|_F^2.
	\end{equation}
	Then the sequence $\{\mathbf{Q}_k,\mathcal{X}_k,\mathcal{E}_k,\mathcal{Z}_k, \mu_k\}$ generated in Algorithm~\ref{Algorithm_1} with Eq.~(\ref{Update_Q_MO}) satisfies the following:
	\begin{equation}
	\begin{aligned}
		&L(\mathbf{Q}_{k},\mathcal{X}_{k},\mathcal{E}_{k},\mathcal{Z}_{k},\mu_{k}) \geq L(\mathbf{Q}_{k+1},\mathcal{X}_{k+1},\mathcal{E}_{k+1},\mathcal{Z}_{k+1},\mu_{k+1}) \\
		&\qquad\qquad + \frac{\mu_k}{2}\left\| \mathcal{E}_k - \mathcal{E}_{k+1} \right\|_F^2 + \left( \frac{\mu_k}{2} - \frac{\mu_{k+1} + \mu_{k}}{2\mu_{k}^2} C_L \right)\left\| \mathcal{X}_k - \mathcal{X}_{k+1} \right\|_F^2.
	\end{aligned}
	\end{equation}
	The function value of Eq.~(\ref{THM_5_ALF}) decreases monotonically after each iteration as long as $\mu \geq \sqrt{(\rho+1) C_L}$, where $\rho$ is defined in Eq.~(\ref{Update_mu}) and $C_L$ is a constant w.r.t $\mathcal{X}$. 
	By the monotone bounded convergence theorem, Algorithm~\ref{Algorithm_1} is convergent.
\end{theorem}

\subsection{Complexity Analysis}\label{Sec:Complexity}

The computational complexity of VMTQN in Eq.~(\ref{Update_Q}) is $\mathcal{O}\left(r n_1 n_2 n_3 \right)$, where $r$ denotes the number of columns of $\mathbf{Q}\in\mathbb{R}^{n_3\times r}$. And the complexity of MOTQN in Eq.~(\ref{Update_Q_MO}) is $\mathcal{O}\left((n_1 n_2+n_3) n_3^2 \right)$. As for TNN based method~\cite{excel_10,excel_20,excel_1_2,LuIJCAI2018}, they use Fourier transform and have a complexity of $\mathcal{O}\left( n_1 n_2 n_3 \log n_3 \right)$. As can be seen, if $r < \log n_3$, VMTQN can be more efficient than the other two methods. Otherwise, we should use a larger $K$ to control the overall calculation speed.

However, when solving our two methods or TNN based method, the most time-consuming part is in the SVD operator of each iteration, which is corresponding to Eqs.~(\ref{Update_X})-(\ref{Update_mu}). In this part, VMTQN based method has a complexity of $\mathcal{O}\left( r n_1 n_2 \min\{n_1,n_2\}  \right)$ while MOTQN and TNN based methods has a complexity of $\mathcal{O}\left( n_3 n_1 n_2 \min\{n_1,n_2\}  \right)$. That is to say, as long as $r \ll n_3$, VMTQN based method is usually more efficient than the other two methods.

\subsection{Performance Analysis}
Considering the low rank tensor recovery models in Eqs.~(\ref{tc_bilevel_model}) and~(\ref{TC_MOTQN}), $\Omega$ is an index set consisting of the indices $\{(i,j,k)\}$ which can be observed, and the orthogonal projection operator $\mathcal{P}_\Omega$ is defined as $\mathcal{P}_\Omega(\mathcal{X}_{ijk}) = \mathcal{X}_{ijk}$ if $(i,j,k)\in\Omega$ and $0$ otherwise. In this part, we discuss at least how many observation samples $|\Omega|$ are needed to recover the ground-truth. In fact, $\mathbf{Q}^*$ obtained from the convergence of Algorithm~\ref{Algorithm_1} has a decisive effect on the number of observation samples needed, since the optimal solution satisfies the KKT conditions under $\mathbf{Q}^*$. That is to say, we only need to analyze the performance guarantee in the case of fixed $\mathbf{Q}$.

With a fixed $\mathbf{Q}$, the exact tensor completion guarantee for model~(\ref{low_tensor_q_rank_model_with_norm}) is shown in Theorem~\ref{Thm_5}. Lu et al~\cite{lu2019cvpr} also have similar conclusions.
\begin{theorem}\label{Thm_5} Given a fixed orthogonal matrix $\mathbf{Q}\in\mathbb{R}^{n_3\times n_3}$ and $\Omega \sim Ber(p)$, assume that tensor $\mathcal{X}\in\mathbb{R}^{n_1\times n_2\times n_3}$~($n_1\geq n_2$) has a low tensor Q-rank structure and $\operatorname{rank}_{Q}(\mathcal{X}) = R$. If $|\Omega| \geq \mathcal{O}(\mu Rn_1 \log(n_1 n_3))$, then $\mathcal{X}$ is the unique solution to Eq.(\ref{low_tensor_q_rank_model_with_norm}) with high probability, where $\Psi$ is replaced by $\mathcal{P}_\Omega$, and $\mu$ is the corresponding incoherence parameter~(See Supplementary Materials).
\end{theorem}

Through the proof of~\cite{lu2019cvpr} and~\cite{LuIJCAI2018}, the sampling rate $p$ should be proportional to $\max\{\|\mathcal{P}_{\mathcal{T}}(\mathfrak{e}_{ijk})\|_F^2\}$.~(The definition of projection operators $\mathcal{P}_{\mathcal{T}}$ and $\mathfrak{e}_{ijk}$ can be found in~\cite{excel_1_2,LuIJCAI2018} or in Supplementary materials, where $\mathcal{T}$ is the singular space of the ground-truth.) 
The projection of $\mathfrak{e}_{ijk}$ onto subspace $\mathcal{T}$ is greatly influenced by the dimension. Obviously, when $\mathcal{T}$ is the whole space, $\left\|\mathcal{P}_{\mathcal{T}_Q}(\mathfrak{e}_{ijk})\right\|_F^2 = 1$.
That is to say, a small dimension of $\mathcal{T}_Q$ may lead to a small $\max_{ijk}\{\left\|\mathcal{P}_{\mathcal{T}_Q}(\mathfrak{e}_{ijk})\right\|_F^2\}$.

\textbf{Proposition~15} in~\cite{LuIJCAI2018} also implies that for any $\Delta \in \mathcal{T}$, we need to have $\mathcal{P}_\Omega(\Delta) = 0 \Leftrightarrow \Delta = 0$. These two conditions indicate that once the spatial dimension of $\mathcal{T}$ is large~($\operatorname{rank}_{Q}(\mathcal{X}) = R$ is large), a larger sampling rate $p$ is needed. And \textbf{Figure~3} in~\cite{LuIJCAI2018} verifies the rationality of this deduction by experiment. 

In fact, the smoothness of data along the third dimension has a great influence on the {Dimension of Freedom}~(DoF) of space $\mathcal{T}$. Non-smooth change along the third dimension is likely to increase the spatial dimension of $\mathcal{T}$ under the Fourier basis vectors, which makes the TNN based methods ineffective. Our experiments on CIFAR-10~(Table~\ref{PSNR_1_2}) confirm this conclusion.

As for the models~(\ref{Eq:VMTQN_TC}) and~(\ref{Eq:MOTQN_TC}) with adaptive $\mathbf{Q}$, our motivation is to find a better $\mathbf{Q}$ in order to make $\operatorname{rank}_{Q}(\mathcal{X}) = R$ smaller and make the spatial dimension of corresponding $\mathcal{T}_Q$ as small as possible, where $\mathcal{T}_Q$ is the singular space of the ground-truth under $\mathbf{Q}$. In other words, for more complex data with non-smoothness along the third dimension, the adaptive $\mathbf{Q}$ may reduce the dimension of $\mathcal{T}_Q$ and make $\max\{\|\mathcal{P}_{\mathcal{T}_Q}(\mathfrak{e}_{ijk})\|_F^2\}$ smaller than $\max\{\|\mathcal{P}_{\mathcal{T}}(\mathfrak{e}_{ijk})\|_F^2\}$, leading to a lower bound for the sampling rate $p$.

\section{Experiments}

In this section, we conduct numerical experiments to evaluate our proposed models~(\ref{Eq:VMTQN_TC}) and~(\ref{Eq:MOTQN_TC}). 
{The platform is Matlab R2018b under Windows 10 on a PC with an Intel i5-7500 CPU and 16GB memory.} The experimental code of most comparison methods come from the released version. As for some methods without released code, we reproduce it in Matlab 2018b strictly according to the algorithm in their respective papers.

Assume that the observed corrupted tensor is $\mathcal{Y}$, and the true tensor is $\mathcal{X}_0 \in \mathbb{R}^{n_1\times n_2\times n_3}$. We represent the recovered tensor (output of the algorithms) as  $\mathcal{X}$, and use Peak Signal-to-Noise Ratio~(PSNR) to measure the reconstruction error:
\begin{equation}
\operatorname{PSNR} = 10\log_{10}{\left( \frac{n_1 n_2 n_3\|\mathcal{X}_0\|^2_\infty}{\|\mathcal{X} - \mathcal{X}_0\|_F^2} \right)}.
\end{equation}
\subsection{Synthetic Experiments}\label{Sec:Synthetic_Experiments}

\begin{figure}[htb]
	\centering
	\includegraphics[width=0.3\columnwidth]{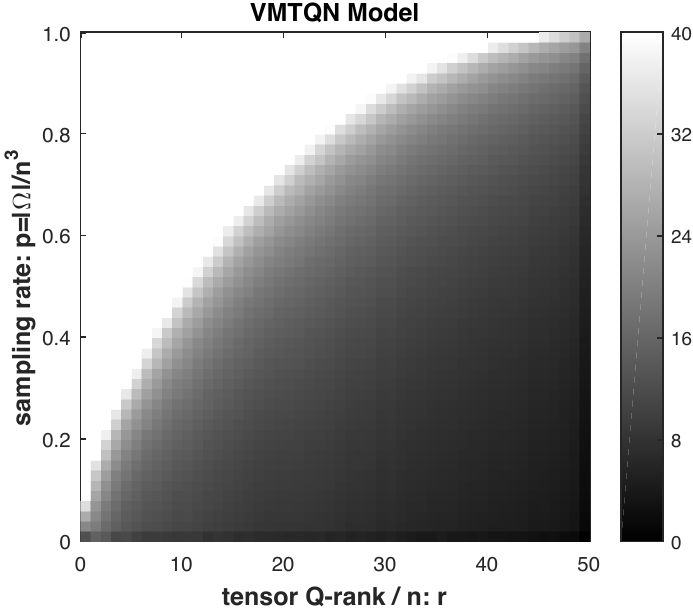}\ 
	\includegraphics[width=0.3\columnwidth]{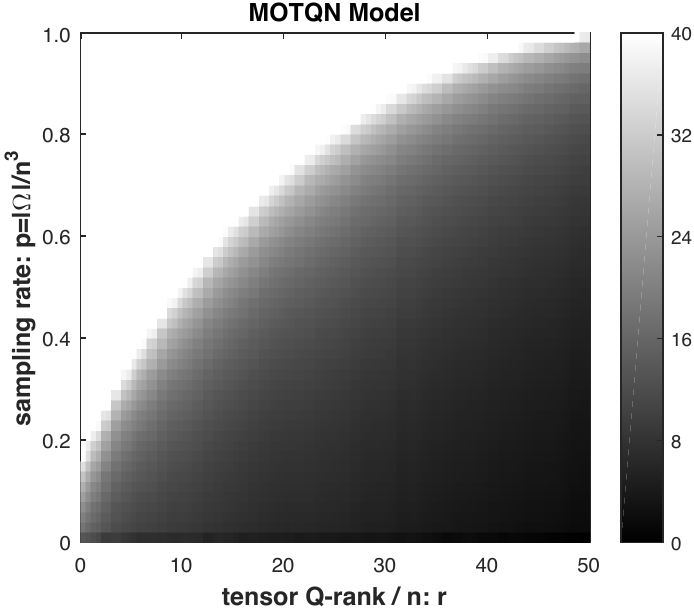}\ 
	\includegraphics[width=0.3\columnwidth]{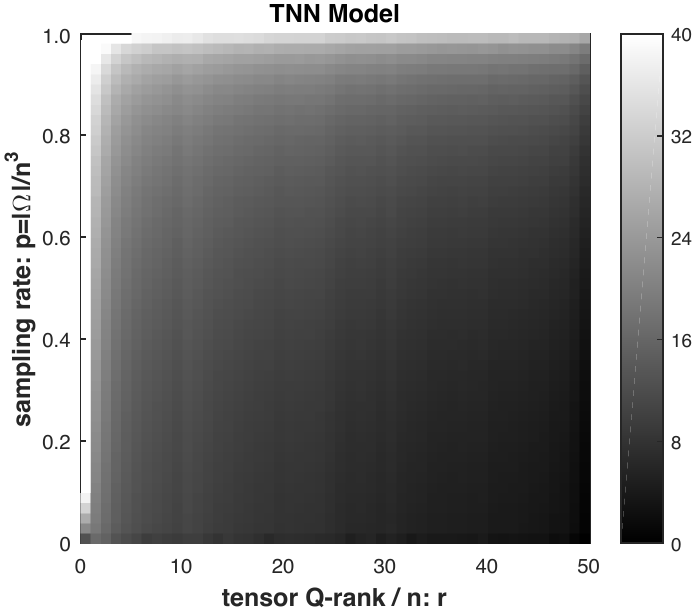}\\
	\caption{The numbers plotted on the above figure are the average PSNRs within 10 random trials. The gray scale reflects the quality of completion results of three different models~(VMTQN, MOTQN, TNN), where $n_1=n_2=n_3=50$ and the white area represents a maximum PSNR of $40$.}\label{gray_square}
\end{figure}

In this part we compare our proposed methods~(named VMTQN model and MOTQN model) with the mainstream algorithm TNN~\cite{excel_10,LuIJCAI2018}.

We examine the completion task with varying tensor Q-rank of tensor $\mathcal{Y}$ and varying sampling rate $p$. Firstly, we generate a random tensor $\mathcal{M}\in\mathbb{R}^{50\times 50\times 50}$, whose entries are independently sampled from an $\mathcal{N}(0,1/50)$ distribution. Actually, the data generated in this way is usually non-smooth along each dimension.
Then we choose $p$ in $[0.01:0.02:0.99]$ and $r$ in $[1:1:50]$, where the column orthonormal matrix $\mathbf{W}\in\mathbb{R}^{50\times r}$ satisfies $\mathbf{W} = \operatorname{PCA}(\mathcal{M},3,r)$. We let $\mathcal{Y} = \mathcal{M}\times_3 \mathbf{W}\times_3 \mathbf{W}^\top$ be the true tensor. 
After that, we create the index set $\Omega$ by using a Bernoulli model to randomly sample a subset from $\{1,\ldots,50\}\times \{1,\ldots,50\}\times \{1,\ldots,50\}$. The sampling rate $p$ is $|\Omega|/ 50^3$.	
For each pair of $(p,r)$, we simulate $10$ times with different random seeds and take the average as the final result. As for the parameters of VMTQN and MOTQN models in Algorithm~\ref{Algorithm_1}, we set $\rho = 1.1$, $\mu_0 = 10^{-4}$, $\mu_{max} = 10^{10}$, and $\epsilon = 10^{-8}$. 

As shown in the upper left corner regions of VMTQN model and MOTQN model in Figure~\ref{gray_square}, Algorithm~\ref{Algorithm_1} can effectively solve our proposed recovery models~(\ref{tc_bilevel_model}) and~(\ref{TC_MOTQN}). The larger tensor Q-rank it is, the larger the sampling rate $p$ is needed, which is consistent with our Performance Analysis in Theorem~\ref{Thm_5}.
By comparing the results of three methods, we can find that TNN has very poor robustness to the data with non-smooth change. And the results of the left and middle images demonstrate our assumptions~(Motivation), which may imply that better low rank structure leads to better recovery.

\subsection{Real-World Datasets}
In this part we compare our proposed method with TNN~\cite{LuIJCAI2018} with Fourier matrix, TTNN~\cite{song2019robust} with wavelet matrix, TNN-C~\cite{xu2019fast} with cosine matrix, F-TNN~\cite{jiang2020framelet} with framelet matrix, SiLRTC~\cite{excel_1_1}, Tmac~\cite{excel_38}, and Latent Trace Norm~\cite{Latent_2013}. We validate our algorithm on three datasets:
(1) CIFAR-10\footnote{ \url{http://www.cs.toronto.edu/~kriz/cifar.html}.}; 
(2) COIL-20\footnote{ \url{http://www.cs.columbia.edu/CAVE/software/softlib/coil-20.php}.}; 
(3) HMDB51\footnote{$  $\url{http://serre-lab.clps.brown.edu/resource/hmdb-a-large-human-motion-database/}.}.
We set $\rho = 1.1$, $\mu_0 = 10^{-4}$, $\mu_{max} = 10^{10}$, $\epsilon = 10^{-8}$, and $K=1$ in our methods. 
{As for TNN, SiLRTC, Tmac, F-TNN, and Latent Trace Norm, we use the default settings as in their released code, e.g., Lu et al.\footnote{\url{https://github.com/canyilu/LibADMM}} and Tomioka et al.\footnote{\url{https://https://github.com/ryotat/tensor}}. For TTNN and TNN-C of unreleased code, we implement their algorithms in MATLAB strictly according to the corresponding papers.}

\begin{table}
	\caption{{Comparisons of PSNR results on CIFAR images with different sampling rates. \textbf{Top:} experiments on the case  $\mathcal{Y}_1\in\mathbb{R}^{32\times 32\times 3000}$. \textbf{Bottom:} experiments on the case $\mathcal{Y}_2\in\mathbb{R}^{32\times 32\times 10000}$.}}
	\label{PSNR_1_2}
	\centering
	\begin{tabular}{lllllll}
		\toprule
		Sampling Rate $p$    & 0.1 & 0.2 & 0.3 & 0.4 & 0.5 & 0.6 \\
		\midrule
		TQN with Random $\mathbf{Q}$ &10.86	&15.47		&18.09		&20.20		&22.30		&24.49 \\
		TQN with Oracle $\mathbf{Q}$~(ideal) &25.39	&30.85		&39.43		&\textbf{109.52}	&\textbf{$>$200}		&\textbf{$>$200} \\
		\midrule
		VMTQN~(Ours) &\textbf{18.83}		&\textbf{21.10}		&\textbf{22.89}		&\textbf{24.56}		&\textbf{26.26}		&\textbf{28.07} \\
		TNN~(Fourier)~\cite{LuIJCAI2018} &9.84		&12.73		&15.68		&18.71		&21.60		&24.26 \\
		TNN-C~(cosine)~\cite{xu2019fast} &9.63		&11.92		&15.17		&18.45	&22.09		&23.95 \\
		TTNN~(wavelet)~\cite{song2019robust} &8.97		&13.08		&17.19		&19.26		&23.13		&25.67 \\
		F-TNN~(framelet)~\cite{jiang2020framelet} &8.84		&11.95		&16.56		&20.61		&23.77		& 26.02 \\
		Tmac~\cite{excel_38} &17.81		&19.29		&23.06		&24.89		&25.74		&27.46 \\
		SiLRTC~\cite{excel_1_1}  &16.87		&20.04		&21.99		&23.80		&25.62		&27.57  \\
		\bottomrule
	\end{tabular}

	\begin{tabular}{lllllll}
		\toprule
		Sampling Rate $p$    & 0.1 & 0.2 & 0.3 & 0.4 & 0.5 & 0.6 \\
		\midrule
		TQN with Random $\mathbf{Q}$ &10.84	&15.45	&18.06	&20.19	&22.29	&24.48 \\
		TQN with Oracle $\mathbf{Q}$~(ideal) &45.75	&\textbf{$>$200}		&\textbf{$>$200}		&\textbf{$>$200}		&\textbf{$>$200}		&\textbf{$>$200} \\
		\midrule
		VMTQN~(Ours) &\textbf{19.06}		&\textbf{21.43}		&\textbf{23.27}		&\textbf{24.97}		&\textbf{26.65}		&\textbf{28.42} \\
		TNN~(Fourier)~\cite{LuIJCAI2018} &8.18		&10.10		&12.19		&14.63		&17.59		&21.20 \\
		TNN-C~(cosine)~\cite{xu2019fast} & 8.12 & 9.95 & 11.80 & 13.62 & 18.07 & 22.10 \\
		TTNN~(wavelet)~\cite{song2019robust} & 9.01 & 10.80 & 13.27 & 15.88 & 20.21 & 24.04 \\
		F-TNN~(framelet)~\cite{jiang2020framelet} & 9.17& 11.06 & 15.10 & 17.44 & 20.85 & 23.77 \\
		Tmac~\cite{excel_38} & 12.91 & 18.49 & 22.97& 25.25 & 27.06 & 27.97\\
		SiLRTC~\cite{excel_1_1}  &14.02		&19.65		&22.44		&24.38		&26.21		&28.12  \\
		\bottomrule
	\end{tabular}
\end{table}

\subsubsection{Influences of $\mathbf{Q}$} Corresponding to our motivation, we use a Random orthogonal matrix and an Oracle matrix~({the matrix of right singular vectors} of the ground-truth unfolding matrix) to test the influence of $\mathbf{Q}$. The results of TQN models with different orthogonal matrix in Tables~\ref{PSNR_1_2} and~\ref{PSNR_3_4} show that $\mathbf{Q}$ play an important role in tensor recovery. Comparing with Random $\mathbf{Q}$ case, our Algorithm~\ref{Algorithm_1} is effective for searching a better $\mathbf{Q}$. Table~\ref{PSNR_1_2} also shows that a proper $\mathbf{Q}$ may make recover the ground-truth more easily. 
For example, with sampling rate $p\geq 0.2$ on $10000$ images, an Oracle matrix $\mathbf{Q}$ can lead to an ``exact'' recovery~($\text{PSNR} >200$).

\subsubsection{CIFAR-10}
We consider the worst case for TNN based methods that there is almost no smoothness along the third dimension of the data. We randomly selected 3000 and 10000 images from one batch of CIFAR-10~\cite{CIFAR} as our true tensors $\mathcal{Y}_1\in\mathbb{R}^{32\times 32\times 3000}$ and $\mathcal{Y}_2\in\mathbb{R}^{32\times 32\times 10000}$, respectively. Then we solve the model~(\ref{Eq:VMTQN_TC}) with our proposed Algorithm~\ref{Algorithm_1}. The results are shown in Table~\ref{PSNR_1_2}. Note that in the latter case $r = n_1n_2 \ll n_3$ holds, MOTQN model has high computational complexity. Thus we will not compare it in this part.

\begin{figure}[t]
	\centering
	\includegraphics[width=0.45\columnwidth]{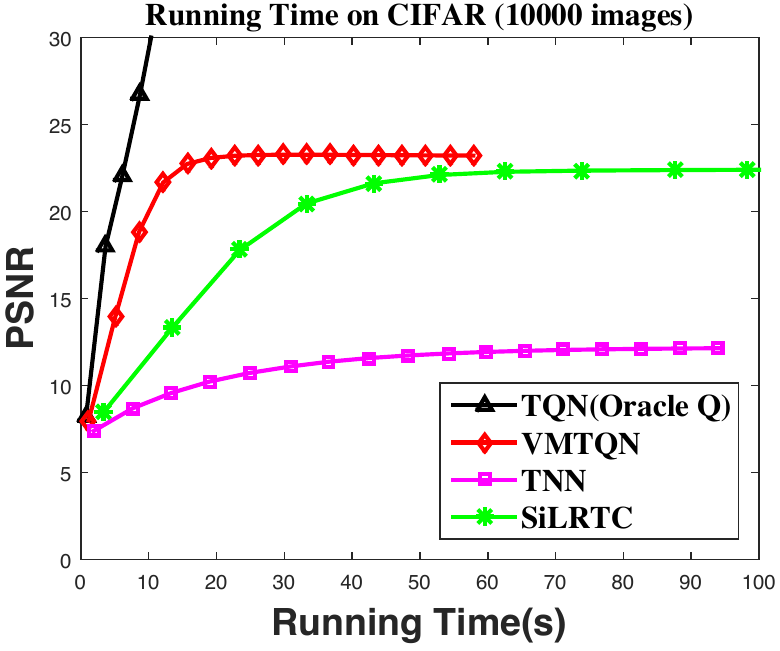}
	\includegraphics[width=0.45\columnwidth]{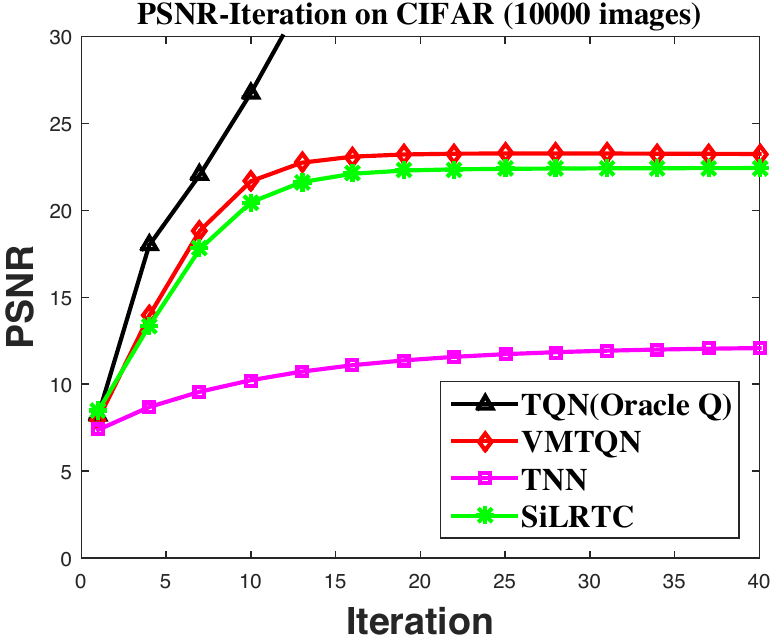}
	\caption{Running time comparisons of different methods, where $\mathcal{Y}\in\mathbb{R}^{32\times 32\times 10000}$ and sampling rate $p=0.3$.}\label{Runningtime_cifar}
\end{figure}

Table~\ref{PSNR_1_2} verifies our hypothesis that TNN regularization performs badly on data with non-smooth change along the third dimension. Our VMTQN method is obviously better than the other methods in the case of low sampling rate. 
Moreover, by comparing the two groups of experiments, we can see that VMTQN, TMac, and SiLRTC perform better in $\mathcal{Y}_2$. This may be due to that increasing the data volume will make the principal components more significant. 
Meanwhile, in the methods of Fourier matrix, cosine matrix and wavelet matrix, they almost have no recovery effect when the sampling rate $p$ is lower. This indicates that these specified projection bases can not learn the data features in the case of poor continuity and insufficient sampling.

The above analyses confirm that our proposed regularization are data-dependent and can lead to a better low rank structure which makes recover easily.

\subsubsection{Running time on CIFAR}
As shown in Figure~\ref{Runningtime_cifar}, we test the running times of different models. The two figures indicate that, when $n_3\gg n_1n_2$, our VMTQN model has higher computational efficiency in each iteration and better accuracy than TNN and SiLRTC. As mentioned in our previous complexity analysis, VMTQN method has a great speed advantage in this case. 
Moreover, for the case $n_3 < n_1n_2$, Figure~\ref{Influence_of_r} implies that setting $r<n_1n_2$ can balance computational efficiency and recovery accuracy.

\begin{figure}[t]
	\centering
	\includegraphics[width=0.6\columnwidth]{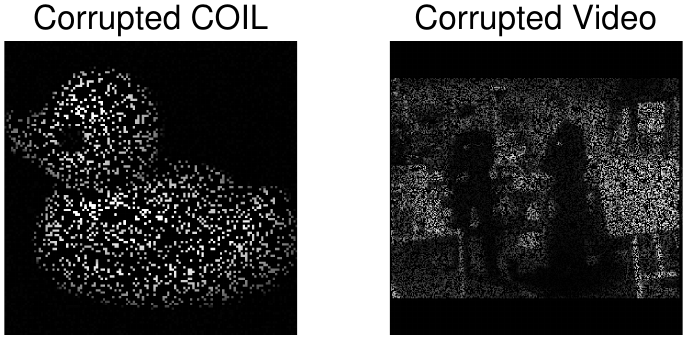}
	\caption{{Examples of the corrupted data in our completion tasks. The left figure is from COIL dataset while the right figure is from the short video. The sampling rate is $p=0.2$ in the left and $p=0.5$ in the right.}}\label{corrupted_coil_and_video}
\end{figure}

\begin{figure}[!htb]
	\centering
	\includegraphics[width=0.65\columnwidth]{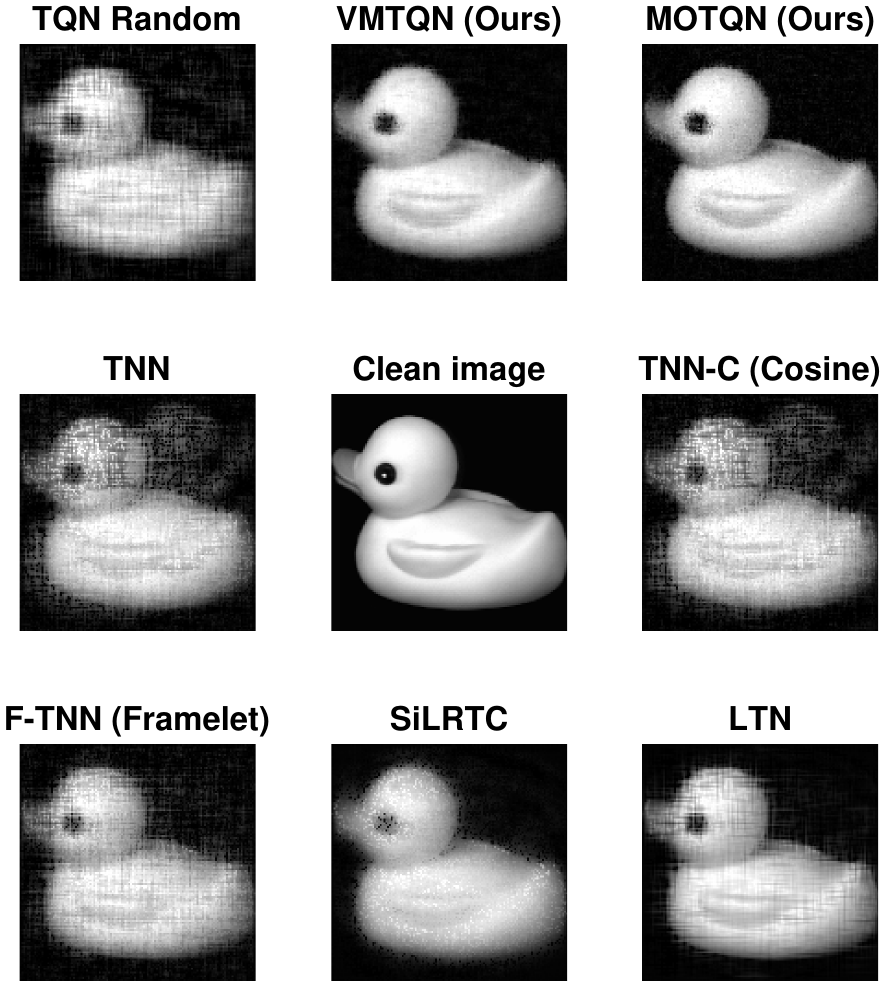}
	\caption{Examples of COIL completion results. Method names correspond to the top of each figure. The sampling rate $p=0.2$.}\label{COIL_0_2_total}
\end{figure}

\begin{figure}[!htb]
	\centering
	\includegraphics[width=0.75\columnwidth]{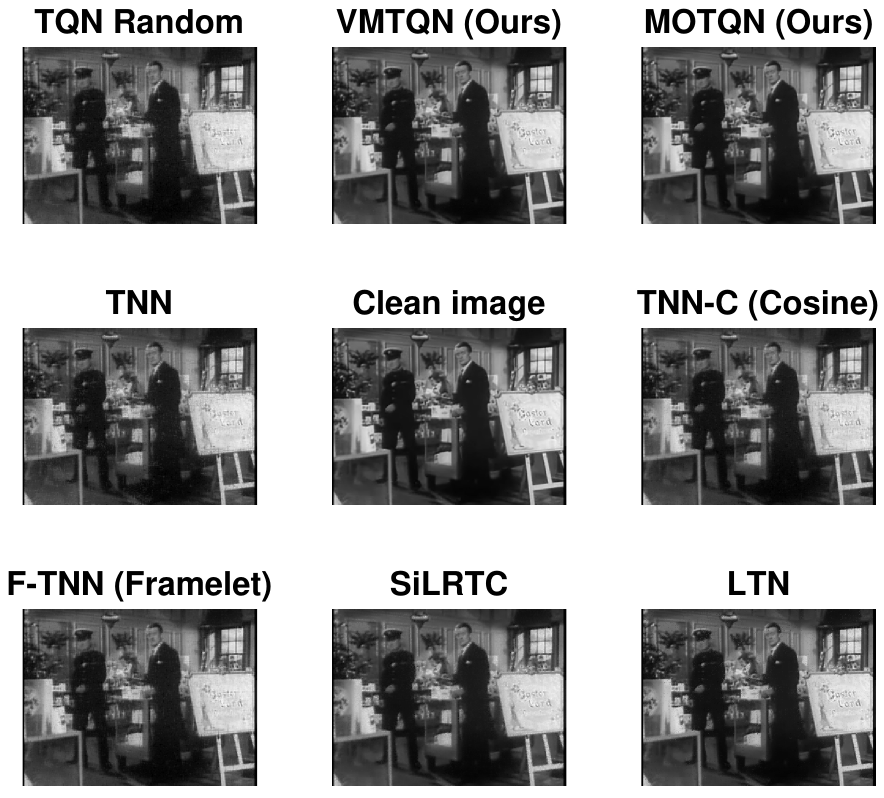}
	\caption{Examples of video inpainting task with sampling rate $p=0.5$.}\label{Video_0_5_total}
\end{figure}

\subsubsection{COIL-20 and Short Video from HMDB51}
COIL-20~\cite{COIL} contains 1440 images of 20 objects which are taken from different angles. The size of each image is processed as $128\times128$, which means $\mathcal{Y}\in\mathbb{R}^{128\times 128\times 1440}$. The upper part of Table~\ref{PSNR_3_4} shows the results of the numerical experiments.
We select a background-changing video from HMDB51~\cite{HMDB} for the video inpainting task, where $\mathcal{Y}\in\mathbb{R}^{240\times 320\times 146}$. Figure~\ref{Intro} shows some frames of this video. The lower part of Table~\ref{PSNR_3_4} shows the results. 
And Figures~\ref{corrupted_coil_and_video},~\ref{COIL_0_2_total} and~\ref{Video_0_5_total} are the the experimental results of COIL-20 and Short Video from HMDB51, respectively.

\begin{table}
	\caption{Comparisons of PSNR results on COIL images and video inpainting with different sampling rates. \textbf{Up:} the COIL dataset with $\mathcal{Y}\in\mathbb{R}^{128\times 128\times 1440}$. \textbf{Down:} a short video from HMDB51 with $\mathcal{Y}\in\mathbb{R}^{240\times 320\times 126}$.}
	\label{PSNR_3_4}
	\centering
	\begin{tabular}{lllllll}
		\toprule
		Sampling Rate $p$    & 0.1 & 0.2 & 0.3 & 0.4 & 0.5 & 0.6 \\
		\midrule
		TQN with Random $\mathbf{Q}$ &16.05	&20.07		&23.02		&25.57		&27.95		&30.34 \\
		TQN with Oracle $\mathbf{Q}$~(ideal) &22.97	&25.32		&27.18		&28.90		&30.68		&32.51 \\
		\midrule
		VMTQN~(Ours) &\textbf{22.79}		&\textbf{25.34}		&\textbf{27.29}		&\textbf{29.08}		&\textbf{30.86}		&\textbf{32.74} \\
		MOTQN~(Ours) &\textbf{21.91} &\textbf{25.41} &\textbf{27.86} &\textbf{30.13} &\textbf{31.79} &\textbf{33.64} \\
		TNN~(Fourier)~\cite{LuIJCAI2018} &19.20		&22.08		&24.45		&26.61		&28.72		&30.91 \\
		TNN-C~(cosine)~\cite{xu2019fast} &19.02 &22.11 & 24.23 &27.04 &28.95 &30.97\\
		TTNN~(wavelet)~\cite{song2019robust} &18.15 & 21.42 &24.47 & 26.93 & 29.11 & 31.10\\
		F-TNN~(framelet)~\cite{jiang2020framelet}  &17.62 &20.58 &22.87 & 24.67 & 27.41 &29.90\\
		Tmac~\cite{excel_38} &19.04 & 22.48 &24.97 & 26.70&27.91 & 28.86\\
		SiLRTC~\cite{excel_1_1}  &18.87		&21.80		&23.89		&25.67		&27.37		&29.14  \\
		Latent Trace Norm~\cite{Latent_2013} &19.09		&22.98		&25.75		&28.11		&30.40		&32.42 \\
		\bottomrule
	\end{tabular}
	
	\begin{tabular}{lllllll}
		\toprule
		Sampling Rate $p$    & 0.1 & 0.2 & 0.3 & 0.4 & 0.5 & 0.6 \\
		\midrule
		TQN with Random $\mathbf{Q}$ &18.85	&22.76		&25.87		&28.73		&31.55		&34.48 \\
		TQN with Oracle $\mathbf{Q}$~(ideal) &23.44	&27.61		&31.37		&35.11		&38.92		&42.74 \\
		\midrule
		VMTQN~(Ours) &\textbf{23.97}		&\textbf{28.09}		&\textbf{31.76}		&\textbf{35.33}		&\textbf{39.06}		&\textbf{42.87} \\
		MOTQN~(Ours) &\textbf{24.10} & \textbf{27.88}&\textbf{32.24} &\textbf{35.19} &\textbf{39.28} &\textbf{42.65}\\
		TNN~(Fourier)~\cite{LuIJCAI2018} &22.40		&25.58		&28.28		&30.88		&33.55		&36.41 \\
		TNN-C~(cosine)~\cite{xu2019fast}  &22.15 &25.34 &28.17 & 30.96 &33.51 & 36.62\\
		TTNN~(wavelet)~\cite{song2019robust} &19.80 & 21.95 & 24.92 & 30.13 & 32.78 & 36.84	 \\
		F-TNN~(framelet)~\cite{jiang2020framelet} &19.01 &23.44 &25.94 &29.32 &32.06 &35.13 \\
		Tmac~\cite{excel_38} &18.54 & 22.79 & 26.08 &29.70 &31.17 & 34.26 \\
		SiLRTC~\cite{excel_1_1}  &18.42		&22.33		&25.76		&29.15		&32.59		&36.15  \\
		Latent Trace Norm~\cite{Latent_2013} &18.94		&22.72		&25.65		&28.26		&30.79		&33.48 \\
		\bottomrule
	\end{tabular}
\end{table}

From the two visual figures we can see that, our VMTQN method and MOTQN method perform the best among all comparative methods. Especially when the sampling rate $p=0.2$ in Figure~\ref{COIL_0_2_total}, our methods has significant superiority in visual evaluation. What's more, ``Latent Trace Norm'' based method performs much better than TNN in COIL, which validates our assumption that with the help of data-dependent $\mathbf{V}$ tensor trace norm is much more robust than TNN in processing non-smooth data. 

Overall, both our methods and t-SVD based methods~(e.g., TNN) perform better than the others~(e.g., SiLRTC) on these two datasets.
It is mainly because the definitions of tensor singular value in tSVD based methods can make better use of the tensor internal structure, and this is also the main difference between tensor Q-nuclear norm~(TQN) and sum of the nuclear norm~(SNN). 

Meanwhile, our method is obviously better than the others at all sampling rates, which reflects the superiority of our data dependent $\mathbf{Q}$.

\begin{figure}[t]
	\centering
	\includegraphics[width=0.45\columnwidth]{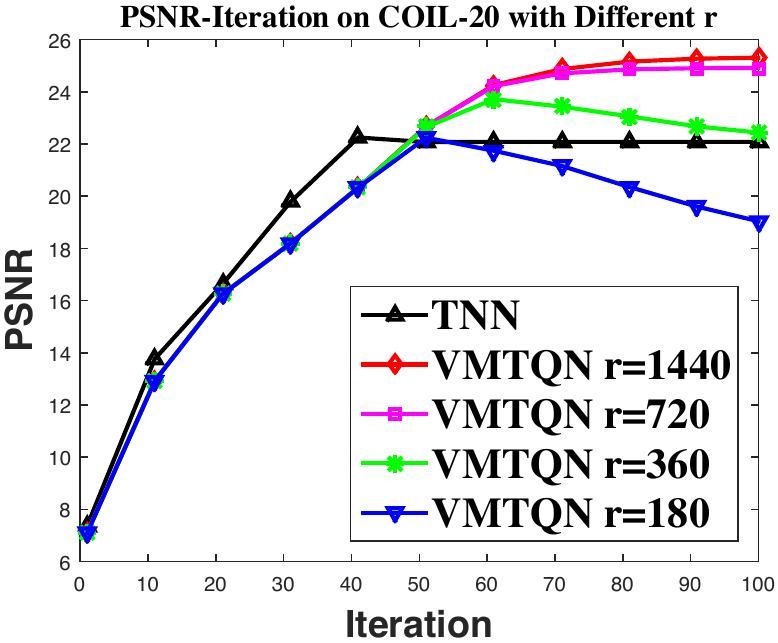}\quad
	\includegraphics[width=0.45\columnwidth]{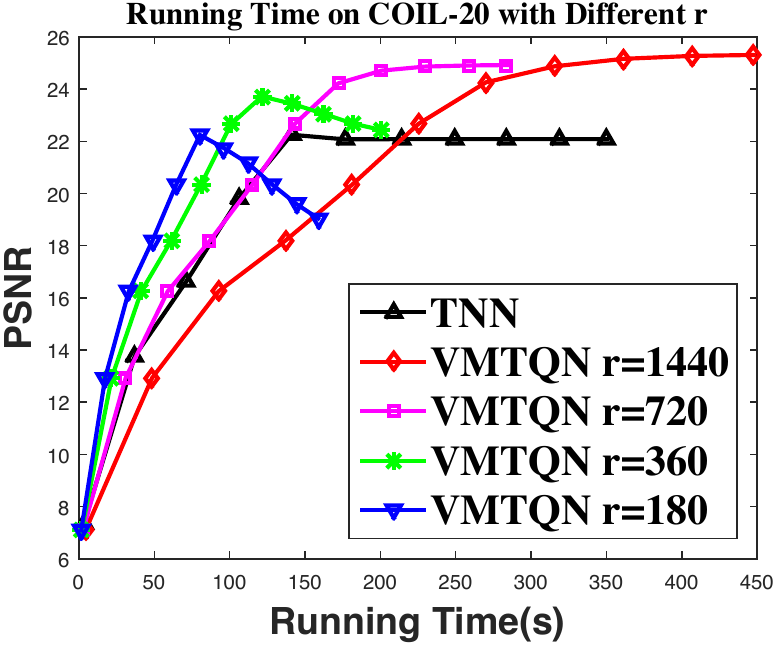}\\
	\includegraphics[width=0.45\columnwidth]{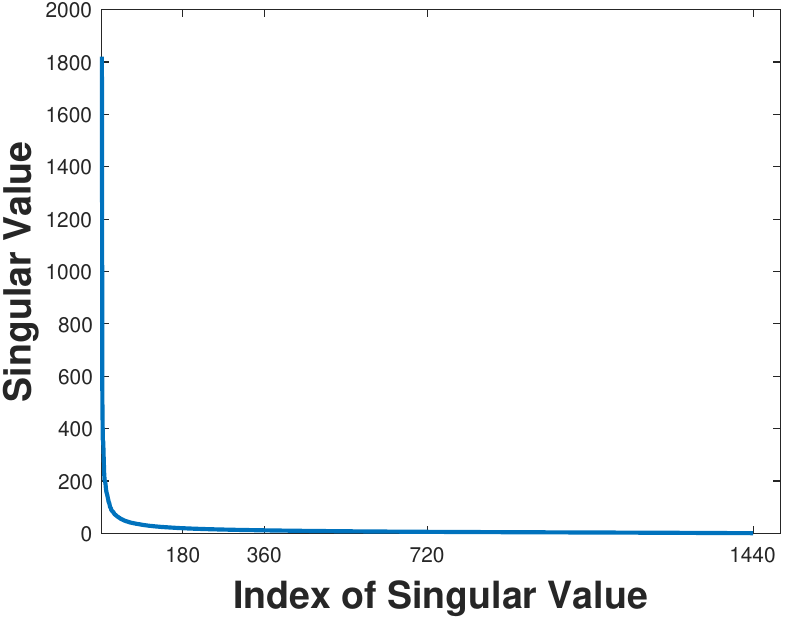}\quad
	\includegraphics[width=0.45\columnwidth]{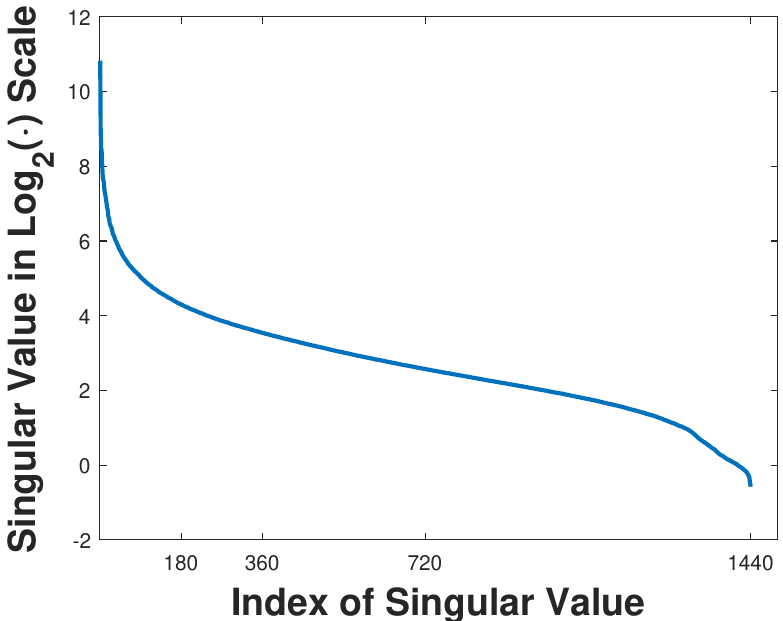}
	\caption{The relations among running times, different $r$, and the singular values of $\mathbf{T}_{(3)}$ on COIL, where $p=0.2$. }\label{Influence_of_r}
\end{figure}

\begin{figure}[h]
	\centering
	\includegraphics[width=0.3\columnwidth]{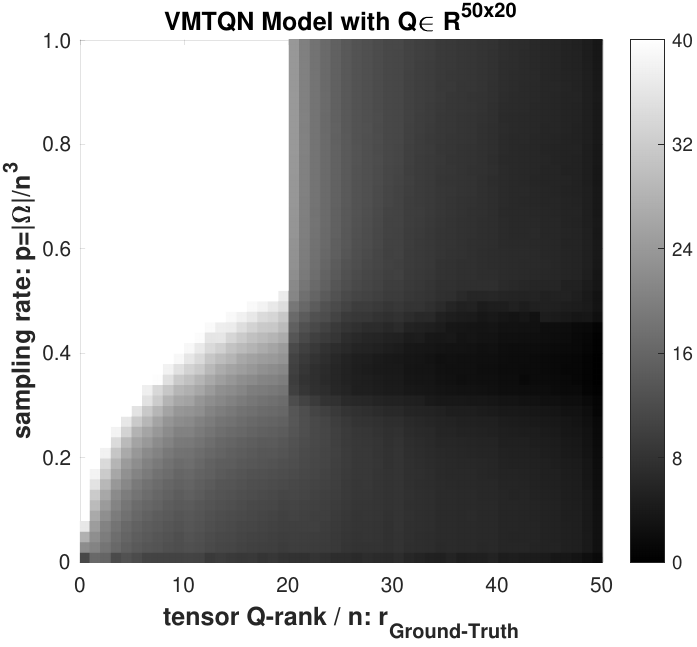}\ 
	\includegraphics[width=0.3\columnwidth]{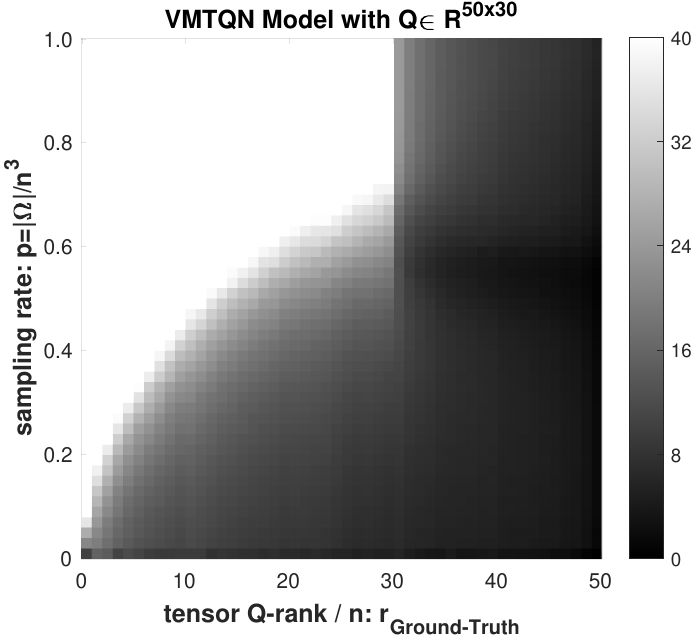}\ 
	\includegraphics[width=0.3\columnwidth]{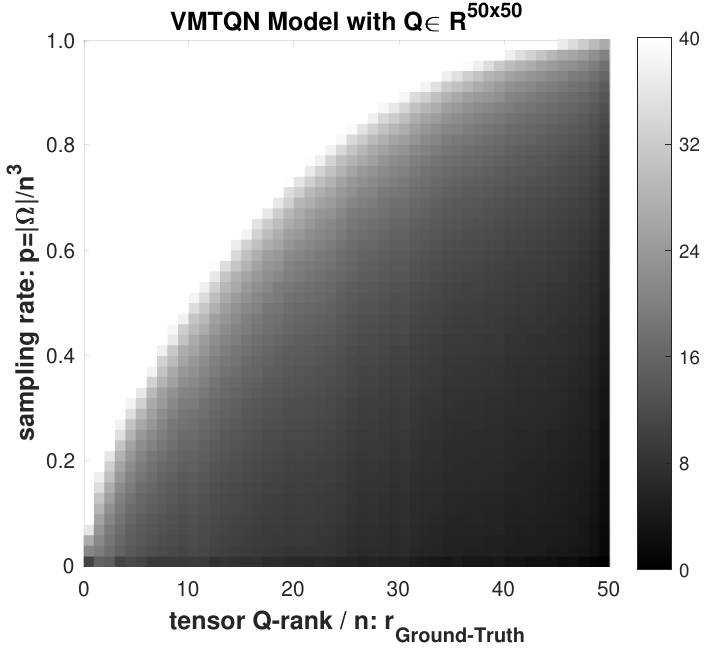}\\
	\caption{The gray scale reflects the quality~(PSNR) of completion results, where $n_1=n_2=n_3=50$ and the white area represents a maximum PSNR of $40$. There are three different sizes of $\mathbf{Q}$ in VMTQN model to show the influences.}\label{Pic:SQ_VMTQN_235}
\end{figure}

\subsubsection{Influence of r in $\mathbf{Q} \in\mathbb{R}^{n_3\times r}$}

Remarks~\ref{Remark_1} and~\ref{Remark_2} imply that $r$ of $\mathbf{Q}\in\mathbb{R}^{n_3\times r}$ in VMTQN denotes the apriori assumption of the subspace dimension of the ground-truth. It means that the dimensions of the frontal slice subspace of the true tensor $\mathcal{T}$~(also as the column subspace of mode-$3$ unfolding matrix $\mathbf{T}_{(3)}$) are no more than $r$. 

Figure~\ref{Influence_of_r} illustrates the relations among running times, different $r$, and the singular values of $\mathbf{T}_{(3)}$. We project the solution $\mathcal{X}_k$~(in Eq.~(\ref{Update_X})) onto the subspace of $\mathbf{Q}_k$, which means $\hat{\mathcal{X}_k} := \mathcal{X}_k\times_3(\mathbf{Q}_k\mathbf{Q}_k^\top)$.
Meanwhile, under different $r$ in $\mathbf{Q} \in\mathbb{R}^{n_3\times r}$, Figure~\ref{Pic:SQ_VMTQN_235} shows the PSNR results of the completion task with varying tensor Q-rank of tensor and varying sampling rate. The settings in Figure~\ref{Pic:SQ_VMTQN_235} are consistent with those in Sec.~\ref{Sec:Synthetic_Experiments}, and only the size of $\mathbf{Q}$ is different.

As shown in the conduct of Figure~\ref{Influence_of_r}, the column subspace of $\mathbf{T}_{(3)}$ is more than 360. If $r\leq 360$, the algorithm will converge to a bad point which only has an $r$-dimensional subspace. 
Therefore, in our previous experiments, we usually set $r=\min\{n_1n_2,n_3\}$ to make sure that $r$ is greater than the true tensor's subspace dimension.
This apriori assumption is commonly used in factorization-based algorithms. What's more, the running time decreases with the decrease of $r$. 
Although $r=1440$ needs more time to converge than TNN, it obtains a better recovery. And a smaller $r$ does speed up the calculation but harms the accuracy.

The results of Figure~\ref{Pic:SQ_VMTQN_235} intuitively reflect the selection criterion of $r$ in VMTQN, that is, $r$ should be larger than the subspace dimension of the true tensor to get the exact recovery. 
{According to the constraint $\mathbf{X}\mathbf{Q}\mathbf{Q}^\top = \mathbf{X}$ in Sec.~\ref{Sec:VMTQN}, if the subspace dimension of the true tensor is larger than $r$, then this constraint can never be satisfied. And and there must be a distance between the output of Algorithm~\ref{Algorithm_1} and the truth tensor, which corresponding to the black areas in the upper right corner of the first two sub-figures.}
From the left two sub-figures we can see that, if the dimension of true tensor is not greater than $r$, the recovery performance is consistent with that in the third sub-figure. 
Combined with the above analyses, $r=\min\{n_1n_2,n_3\}$ can not only save computational efficiency in some cases, but also make the recovery performance of the model in ``the white area'', corresponding to the exact recovery.

\subsection{Smooth Data Experiments}
\begin{figure}
	\begin{minipage}[b]{\linewidth}
		\centering
		\begin{tabular}[b]{llllll}
			\toprule
			Sampling Rate $p$     & 0.2 & 0.3 & 0.4 & 0.5 & 0.6 \\
			\midrule
			TNN~(Fourier)~\cite{LuIJCAI2018}   & 25.64		&	28.08	&	30.43	&	32.82	& 35.36\\
			VMTQN~(Ours) 		&\textbf{26.58}		&\textbf{29.18}		&\textbf{31.69}		&\textbf{34.21}		&\textbf{36.87} \\
			MOTQN~(Ours)  &{26.34} &{28.55} &{30.07} &{32.41} &{36.59} \\
			\bottomrule
		\end{tabular}
	\end{minipage}
	\newline
	\newline
	\begin{minipage}[b]{\linewidth}
	\centering
	\includegraphics[width=\columnwidth]{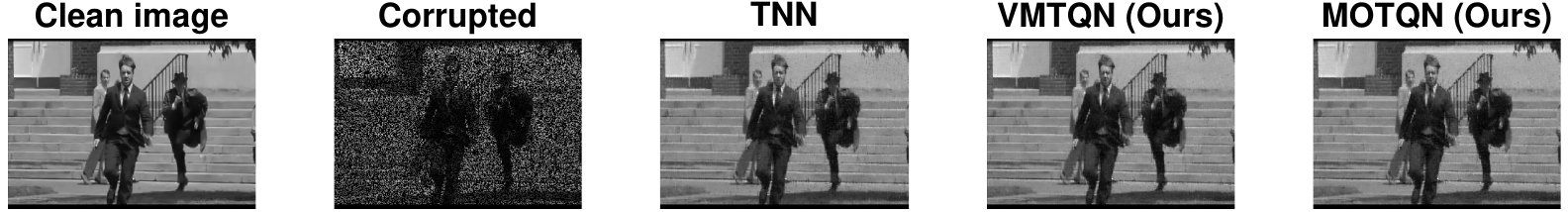}
	\end{minipage}
	\caption{Comparisons of PSNR and visualization results of a smooth video inpainting. \textbf{Up:} PSNR results with different sampling rates. \textbf{Down:} visualization results with the sampling rate $p=0.5$.}
	\label{Pic:Smooth_Data}
\end{figure}

To verify the effectiveness of our proposed methods in smooth data, we select a video from HMDB51 to conduct the experiments, while the background of this video remains unchanged. Figure~\ref{Pic:Smooth_Data} shows the PSNR and visualization results of the video inpainting tasks. Here we only compare TNN based method~\cite{LuIJCAI2018}, since in recent years TNN is considered as a benchmark for handling such smooth data. The results in Figure~\ref{Pic:Smooth_Data} shows that VMTQN method performs best, and with the increase of sampling rate $p$, MOTQN method outperforms TNN based method, which means our proposed methods are still competitive in processing smooth data.

\section{Conclusions}

We analyze the advantages and limitations of the current mainstream low rank regularizers, and then introduce a new definition of data dependent tensor rank named tensor Q-rank. To get a more significant low rank structure w.r.t. $\text{rank}_Q$, we further introduce two explainable selection method of $\mathbf{Q}$ and make $\mathbf{Q}$ to be a learnable variable w.r.t. the data. Specifically, maximizing the variance of singular value distribution leads to VMTQN, while minimizing the value of nuclear norm through manifold optimization leads to MOTQN.
We provide an envelope of our rank function and apply it to the tensor completion problem. By analyzing the proof of exact recovery theorem,we explain why our method may perform better than TNN based methods in non-smooth data~(along the third dimension) with low sampling rates, and conduct experiments to verify our conclusions.

\appendix

\section{Proof of Lemma~\ref{Intro_lemma_1}}
\label{APP_1}

\begin{proof}
	Suppose that $\bar{a} = \frac{1}{n}\sum_{i=1}^{n} a_i$, hence the variance of $\{a_1,\ldots,a_n\}$ can be expressed as $\text{Var}[a_i] = \sum_{i=1}^{n}(a_i - \bar{a})^2$. With $\sum_{i=1}^{n} a_i^2 = C$ holds, we have the following:
	\begin{equation*}
	\begin{aligned}
	\max\ \text{Var}[a_i] 
	\Rightarrow \max\  &\sum_{i=1}^{n}(a_i - \bar{a})^2
	\Rightarrow \max\  \sum_{i=1}^{n}(a_i^2 + \bar{a}^2 - 2 a_i \bar{a})\\
	\Rightarrow \max\  & (\sum_{i=1}^{n}a_i^2) + (\sum_{i=1}^{n}\bar{a}^2) - 2(\sum_{i=1}^{n} a_i \bar{a})\\
	\Rightarrow \max\  & n\bar{a}^2 - 2 \bar{a} (n\bar{a})
	\Rightarrow \max\   -n\bar{a}^2
	\Rightarrow \min\   \bar{a}\quad (\text{due to } a_i \geq 0).
	\end{aligned}
	\end{equation*}
	Moreover, the feasible region of $\{a_1,\ldots,a_n \}$ is an first quadrant Euclidean spherical surface: $\{(a_1,\ldots,a_n)|\sum_{i=1}^{n} a_i^2 = C, a_i\geq 0\}$. Thus the objective function $\bar{a} = \frac{1}{n}\sum_{i=1}^{n} a_i$ is actually a linear hyperplane optimization problem, whose optimal solution contains all intersection of the sphere and each axis, which corresponds to only one non-zero coordinate in $\{a_1,\ldots,a_n\}$.
	\qed
\end{proof}

\section{Proof of Lemma~\ref{Intro_lemma_2}} 
\label{APP_2}

\begin{proof}
	Firstly, $\mathbf{X} = \mathbf{U}\mathbf{\Sigma}\mathbf{V}^\top$ denotes the full Singular Value Decomposition of matrix $\mathbf{X}$ with $\mathbf{U}\in\mathbb{R}^{n_1\times n_1}$, $\mathbf{\Sigma}\in\mathbb{R}^{n_1\times n_2}$, and $\mathbf{V}\in\mathbb{R}^{n_2\times n_2}$. And $\mathbf{P} = \mathbf{V}^\top \mathbf{Q}$ is also an orthogonal matrix, where $\mathbf{P}\in\mathbb{R}^{n_2\times n_2}$. We use $P_{ij}$ to represent the $(i,j)$-th element of matrix $\mathbf{P}$, and use $\mathbf{p}_i$ to represent the $i$-th column of matrix $\mathbf{P}$. Then $\mathbf{X}\mathbf{Q} = \mathbf{U}\mathbf{\Sigma}\mathbf{V}^\top\mathbf{Q} = \mathbf{U}\mathbf{\Sigma}\mathbf{P}$ holds and we have the following:
	\begin{equation}\label{Appendix_L2}
	\|\mathbf{X}\mathbf{Q}\|_{2,1} = \|\mathbf{U}\mathbf{\Sigma}\mathbf{P}\|_{2,1} = \sum_{i=1}^{n_2} \|\mathbf{U}\mathbf{\Sigma} \mathbf{p}_i\|_2 = \sum_{i=1}^{n_2} \|\mathbf{\Sigma} \mathbf{p}_i\|_2.
	\end{equation}
	If $n_1 \geq n_2$, let $\sigma_i = \mathbf{\Sigma}_{ii}$ be the $(i,i)$-th element value of $\mathbf{\Sigma}$ with $i=1,\ldots,n_2$. Or if $n_1 < n_2$, let $\mathbf{\Sigma}' = \begin{pmatrix}
	\mathbf{\Sigma}\\ \mathbf{0}
	\end{pmatrix} \in \mathbb{R}^{n_2\times n_2}$ and $\sigma_i = \mathbf{\Sigma}'_{ii}$ with $i=1,\ldots,n_2$. 
	In this case, $\sum_{i=1}^{n_2} \|\mathbf{\Sigma} \mathbf{p}_i\|_2 = \sum_{i=1}^{n_2} \|\mathbf{\Sigma}' \mathbf{p}_i\|_2$. Thus, we can always get $\{\sigma_1,\ldots,\sigma_{n_2}\}$ and have the equation $\sum_{i=1}^{n_2} \|\mathbf{\Sigma} \mathbf{p}_i\|_2 = \sum_{i=1}^{n_2} \sqrt{\sum_{j=1}^{n_2}(\sigma_j P_{ji})^2}$.

	We then prove that $\mathbf{P} = \mathbf{I}$ optimize the problem~(\ref{Lemma_1_equation}). By using Eq.~(\ref{Appendix_L2}), the objective function can be written as $\sum_{i=1}^{n_2} \|\mathbf{\Sigma} \mathbf{p}_i\|_2$. We give the following deduction:
	\begin{equation*}
	\begin{aligned}
	\sum_{i=1}^{n_2} \|\mathbf{\Sigma} \mathbf{p}_i\|_2 =& \sum_{i=1}^{n_2} \sqrt{\sum_{j=1}^{n_2}(\sigma_j P_{ji})^2}
	\overset{(a)}{=}\sum_{i=1}^{n_2} \sqrt{\sum_{j=1}^{n_2}(\sigma_j P_{ji})^2 \times \sum_{j=1}^{n_2}P_{ji}^2 }\\
	\overset{(b)}{\geq}& \sum_{i=1}^{n_2}  \sum_{j=1}^{n_2} (\sigma_j P_{ji}^2)
	\overset{(c)}{=} \sum_{j=1}^{n_2} \sigma_j \left( \sum_{i=1}^{n_2} P_{ji}^2 \right) 
	\overset{(d)}{=} \sum_{j=1}^{n_2} \sigma_j.
	\end{aligned}
	\end{equation*}
	$(a)$ holds due to that $\mathbf{P}$ is an orthogonal matrix with normalized columns. $(b)$ holds because of Cauchy inequality. $(c)$ holds with exchanging the order of two summations. Finally $(d)$ holds owing to the row normalization of $\mathbf{P}$. Notice that the equality in $(b)$ holds if and only if the two vectors $ (\sigma_1 P_{1i}, \ldots, \sigma_{n_2} P_{n_2 i})$ and $ (P_{1i}, \ldots, P_{n_2 i}) $ are parallel. It can be seen that when $\mathbf{P} = \mathbf{I}$, the condition are satisfied. In other words, $\mathbf{V}^\top\mathbf{Q} = \mathbf{I}$ optimize the problem~(\ref{Lemma_1_equation}), which implies $\mathbf{Q} = \mathbf{V}$.
	\qed
\end{proof}

\newpage
\section{Proof of Theorem~\ref{Theorem_From_Lemma_2}}
\label{APP_3}
\begin{proof}
	We divide $r=\min\{n_1,n_2\}$ into two cases and prove them respectively. {And we use the same notation as in the previous proofs.}
	
	\textbf{(1):} If $n_1 < n_2$ and $r = n_1$, then $\mathbf{U}\in\mathbb{R}^{n_1\times n_1}$, $\mathbf{V} \in\mathbb{R}^{n_2\times n_1}$, and $\mathbf{Q} \in\mathbb{R}^{n_2\times n_1}$. In this case, $\mathbf{\Sigma}\in\mathbb{R}^{n_1\times n_1}$. Let $\mathbf{\Sigma}' = \begin{pmatrix}
	\mathbf{\Sigma}& \mathbf{0}
	\end{pmatrix}\in \mathbb{R}^{n_1\times n_2}$, $\mathbf{V}' = \begin{pmatrix}
	\mathbf{V}& \mathbf{V}_\perp 
	\end{pmatrix} \in \mathbb{R}^{n_2\times n_2}$, and $\mathbf{Q}' = \begin{pmatrix}
	\mathbf{Q}& \mathbf{Q}_\perp 
	\end{pmatrix} \in \mathbb{R}^{n_2\times n_2}$. Note that the constraint $\mathbf{X}\mathbf{Q}\mathbf{Q}^\top = \mathbf{X}$ in Eq.~(\ref{add_THM_EQ}) implies $\mathbf{V}^\top\mathbf{Q}_\perp = \mathbf{0}$ and $\mathbf{V}_\perp^\top \mathbf{Q} = \mathbf{0}$, then we have the following: 
	\begin{equation}
	\|\mathbf{X}\mathbf{Q}\|_{2,1} = \|\mathbf{U}\mathbf{\Sigma}\mathbf{V}^\top\mathbf{Q}\|_{2,1} 
	= \|\mathbf{\Sigma}\mathbf{V}^\top\mathbf{Q}\|_{2,1}
	= \|\mathbf{\Sigma}'\mathbf{V}'^\top\mathbf{Q}'\|_{2,1}.
	\end{equation}
	
	That is to say, minimize $\|\mathbf{X}\mathbf{Q}\|_{2,1}$ w.r.t. $\mathbf{Q}$ in Eq.~(\ref{add_THM_EQ}) is equivalent to minimize $\|\mathbf{\Sigma}'\mathbf{V}'^\top\mathbf{Q}'\|_{2,1}$ w.r.t. $\mathbf{Q}'$ {under the constraints}  $\mathbf{V}^\top\mathbf{Q}_\perp = \mathbf{0}$ and $\mathbf{V}_\perp^\top \mathbf{Q} = \mathbf{0}$.
	By using Lemma~\ref{Intro_lemma_2}, $\mathbf{Q}' = \mathbf{V}'$ minimize the objective function $\|\mathbf{\Sigma}'\mathbf{V}'^\top\mathbf{Q}'\|_{2,1}$, which also satisfies the constraints. In other words, $\mathbf{Q} = \mathbf{V}$ optimize the problem~\ref{add_THM_EQ}.

	\textbf{(2):} If $n_1 \geq n_2$ and $r = n_2$, then $\mathbf{U}\in\mathbb{R}^{n_1\times n_2}$, $\mathbf{V} \in\mathbb{R}^{n_2\times n_2}$, and $\mathbf{Q} \in\mathbb{R}^{n_2\times n_2}$. In this case, we have	
	\begin{equation*}
	\begin{aligned}
	\|\mathbf{X}\mathbf{Q}\|_{2,1} =& \|\mathbf{U}\mathbf{\Sigma}\mathbf{P}\|_{2,1} = \sum_{i=1}^{n_2} \|\mathbf{U}\mathbf{\Sigma} \mathbf{p}_i\|_2
	=&\sum_{i=1}^{n_2} \|\mathbf{\Sigma} \mathbf{p}_i\|_2.
	\end{aligned}
	\end{equation*}
	The remaining proofs are similar to the details in Appendix~\ref{APP_2}. 
	\qed
\end{proof}

\section{Proof of Lemma~\ref{first_and_second_d_of_f}}
\label{APP_4}
\begin{proof}
	Let $g(\tau) = f(\mathbf{Q}(\tau)) = \|\mathcal{X}\|_{Q(\tau),*}$ and $\mathbf{Q}(\tau) \approx \left( \mathbf{I} - \tau \mathbf{A}+ \frac{\tau^2}{2}\mathbf{A}^2 \right)\mathbf{Q}_k $, where $\mathbf{A}$ is defined in Eq.~(\ref{construct_curve}).
	We consider the following approximation:
	\begin{equation}
		g(\tau) = f(\mathbf{Q}(\tau)) \approx g(0) + \left\langle \frac{\partial f(\mathbf{Q}(\tau))}{\partial \mathbf{Q}(\tau)}\Big|_{\tau = 0}, \mathbf{Q}(\tau) - \mathbf{Q}(0) \right\rangle = g(0) + \left\langle  \mathbf{X}_{(3)}^\top \mathbf{H}_{(3)}, \mathbf{Q}(\tau) - \mathbf{Q}(0) \right\rangle,
	\end{equation}
	where $\mathbf{Q}(0) = \mathbf{Q}_k$ and then Eq.~(\ref{gradient}) ensure $\frac{\partial f(\mathbf{Q}(\tau))}{\partial \mathbf{Q}(\tau)}\Big|_{\tau = 0} =  \mathbf{X}_{(3)}^\top \mathbf{H}_{(3)}$. Then we have:
	\begin{equation}
		g(\tau) \approx \left\langle \mathbf{X}_{(3)}^\top \mathbf{H}_{(3)}, \left( \mathbf{I} - \tau \mathbf{A}+ \frac{\tau^2}{2}\mathbf{A}^2 \right)\mathbf{Q}_k  \right\rangle + C_\tau, 
	\end{equation}
	where $C_\tau$ is a constant independent of $\tau$. Then the first and the second order derivatives of $g(\tau)$ evaluated at 0 can be estimated as follows:
	\begin{equation}
	\begin{aligned}
	g'(0)  \approx  \left\langle \mathbf{X}_{(3)}^\top \mathbf{H}_{(3)}, -\mathbf{A} \mathbf{Q}_k \right\rangle, \text{ and }
	g''(0) \approx  \left\langle \mathbf{X}_{(3)}^\top \mathbf{H}_{(3)}, \mathbf{A}^2\mathbf{Q}_k \right\rangle, \\
	\end{aligned}
	\end{equation}
	\qed
\end{proof}


\bibliographystyle{ieeetr}
\bibliography{tqrref}

\end{document}